%% file: root.tex
\documentclass[Afour,sagev,times]{sagej}

\usepackage{moreverb,url}

\usepackage[colorlinks,bookmarksopen,bookmarksnumbered,citecolor=red,urlcolor=red]{hyperref}
\usepackage{algorithmic}
\usepackage{algorithm}
\usepackage{cancel}
\newcommand\BibTeX{{\rmfamily B\kern-.05em \textsc{i\kern-.025em b}\kern-.08em
T\kern-.1667em\lower.7ex\hbox{E}\kern-.125emX}}
\newcommand{\ourrep}{ori\texttt{CORN}}
\newcommand{\ournetwork}{DEF}
\newcommand{\entirename}{\texttt{DEF-}ori\texttt{CORN}}

\def\volumeyear{2024}

\begin{document}

\runninghead{Son et al.}

\title{DEF-oriCORN: efficient 3D scene understanding for robust language-directed manipulation without demonstrations}

\author{Dongwon Son\affilnum{1}, Sanghyeon Son\affilnum{1}, Jaehyung Kim\affilnum{1}, and Beomjoon Kim\affilnum{1}}

\affiliation{\affilnum{1}Korea Advanced Institute of Science and Technology, Daejeon, Republic of Korea}

\corrauth{Beomjoon Kim, beomjoon.kim@kaist.ac.kr}

\begin{abstract}
We present \entirename, a framework for language-directed manipulation tasks. By leveraging a novel object-based scene representation and diffusion-model-based state estimation algorithm, our framework enables efficient and robust manipulation planning in response to verbal commands, even in tightly packed environments with sparse camera views without any demonstrations. Unlike traditional representations, our representation affords efficient collision checking and language grounding. Compared to state-of-the-art baselines, our framework achieves superior estimation and motion planning performance from sparse RGB images and zero-shot generalizes to real-world scenarios with diverse materials, including transparent and reflective objects, despite being trained exclusively in simulation. Our code for data generation, training, inference, and pre-trained weights are publicly available at: \url{https://sites.google.com/view/def-oricorn/home}.
\end{abstract}


\maketitle

\input{intro-bk-v2}
\input{intro_sec1-2}

\input{related_works}

\input{method}

\input{experiment_v2}

\section{Conclusion}
In this work, we propose~\entirename~for object-based scene representation and estimation tailored for language-directed manipulation tasks. Our novel representation \ourrep\ leverages a SO(3)-equivariant network and three decoders to encode oriented shapes into a neural representation, enabling efficient and robust manipulation planning in complex environments. We demonstrate that our method outperforms conventional occupancy-based implicit representations by efficiently grounding language commands to objects and predicting collision without the need for extensive surface point evaluation or mesh conversion. Our integrated estimation and representation framework, \entirename, effectively handles the uncertainty inherent in sparse camera views and partial observability by integrating a diffusion model into the iterative network structure. This enhances the success rate of grasping and motion planning tasks. Future work will focus on refining the estimation process and exploring the potential of our framework in more diverse and dynamic scenarios. Additionally, we plan to explore the integration of task and motion planning algorithms to extend our framework's applicability to long-horizon tasks, enhancing its utility for more comprehensive planning scenarios.


\bibliographystyle{SageV}
\bibliography{references}

\input{appendix}

\end{document}

%% file: intro-bk-v2.tex
\section{Introduction}
\begin{figure}[t]
    \centering
    \includegraphics[width=0.98\linewidth]{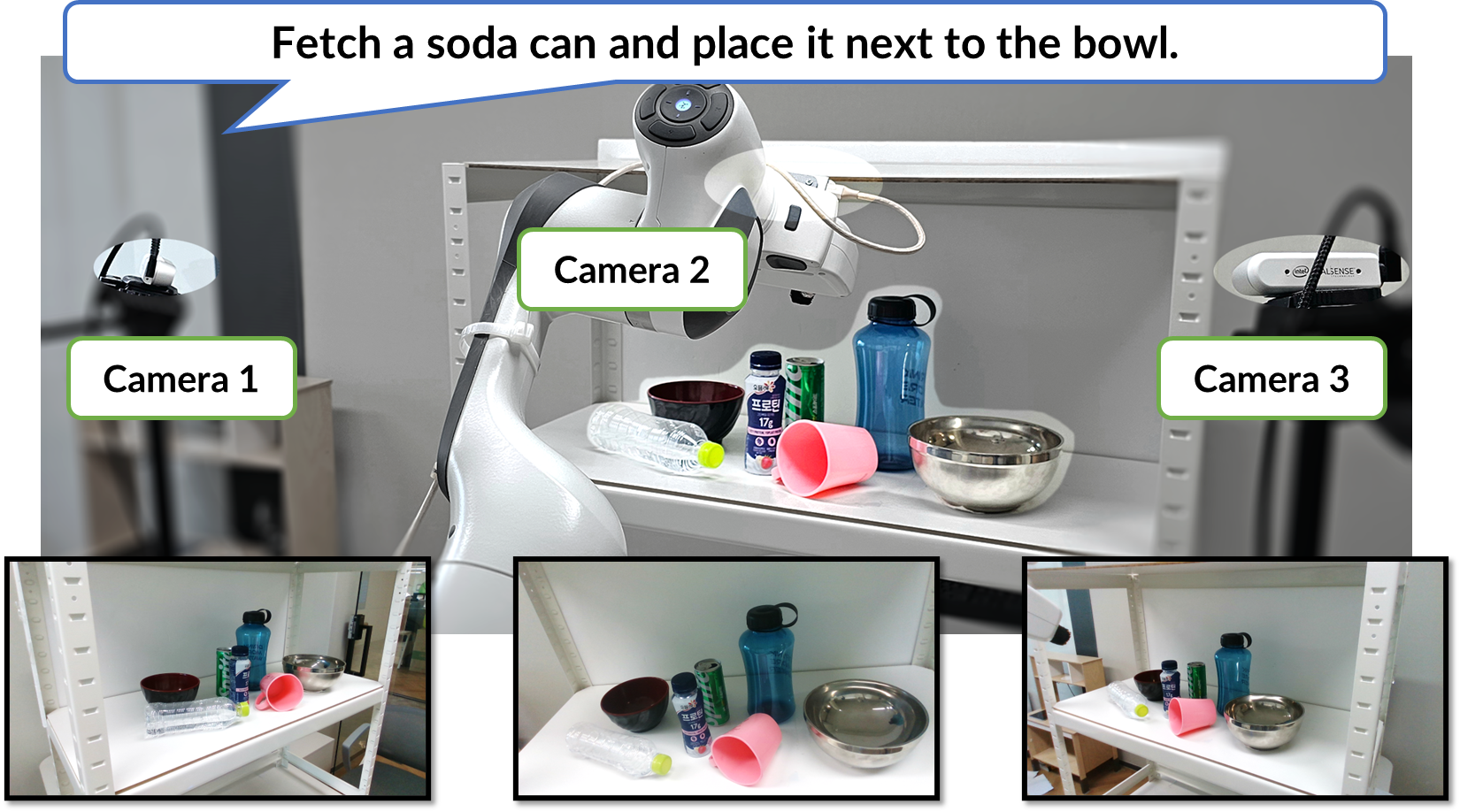}
    \caption{An example problem. We use three RGB cameras, one of which is attached to the end-effector, and the rest are fixed facing the cabinet. Multiple objects including shiny and transparent objects are on the shelf. The robot must ground the language command to objects and plan a collision-free pick-and-place motion to achieve the goal.}
    \label{fig:problem_setup}
\end{figure}

Even with a single glimpse, humans quickly create a mental representation of a scene that enables them to interact with it~\cite{mentalmodels}. We can efficiently estimate 3D shapes and locations of even occluded objects with a sparse set of viewpoints, and achieve language goals such as ``fetch a soda can and place it next to the bowl'', as shown in Figure~\ref{fig:problem_setup}, by planning a manipulation motion that accounts for uncertainty caused by partial observation. Our goal in this paper is to endow robots with such capability.

One of the core challenges in this problem is acquiring an object state representation that encodes the information necessary to ground language commands and plan manipulation motions. A common object state representation is a pair of a 6D pose and an implicit shape function~\cite{OccupancyNetworks,irshad2022centersnap,irshad2022shapo}, but this has limitations. First, 3D orientations are ill-defined for symmetric objects, making estimation under partial observability difficult. Second, implicit shape functions are expensive to use for manipulation due to the need for surface point sampling and mesh creation for collision checking~\cite{rrt,siciliano2008springer}. Third, pose and shape detection often rely on depth sensors, which struggle with transparent or shiny objects. Lastly, grounding language commands to objects using this representation is non-trivial.


\begin{figure*}
    \centering
    \includegraphics[width=0.99\linewidth]{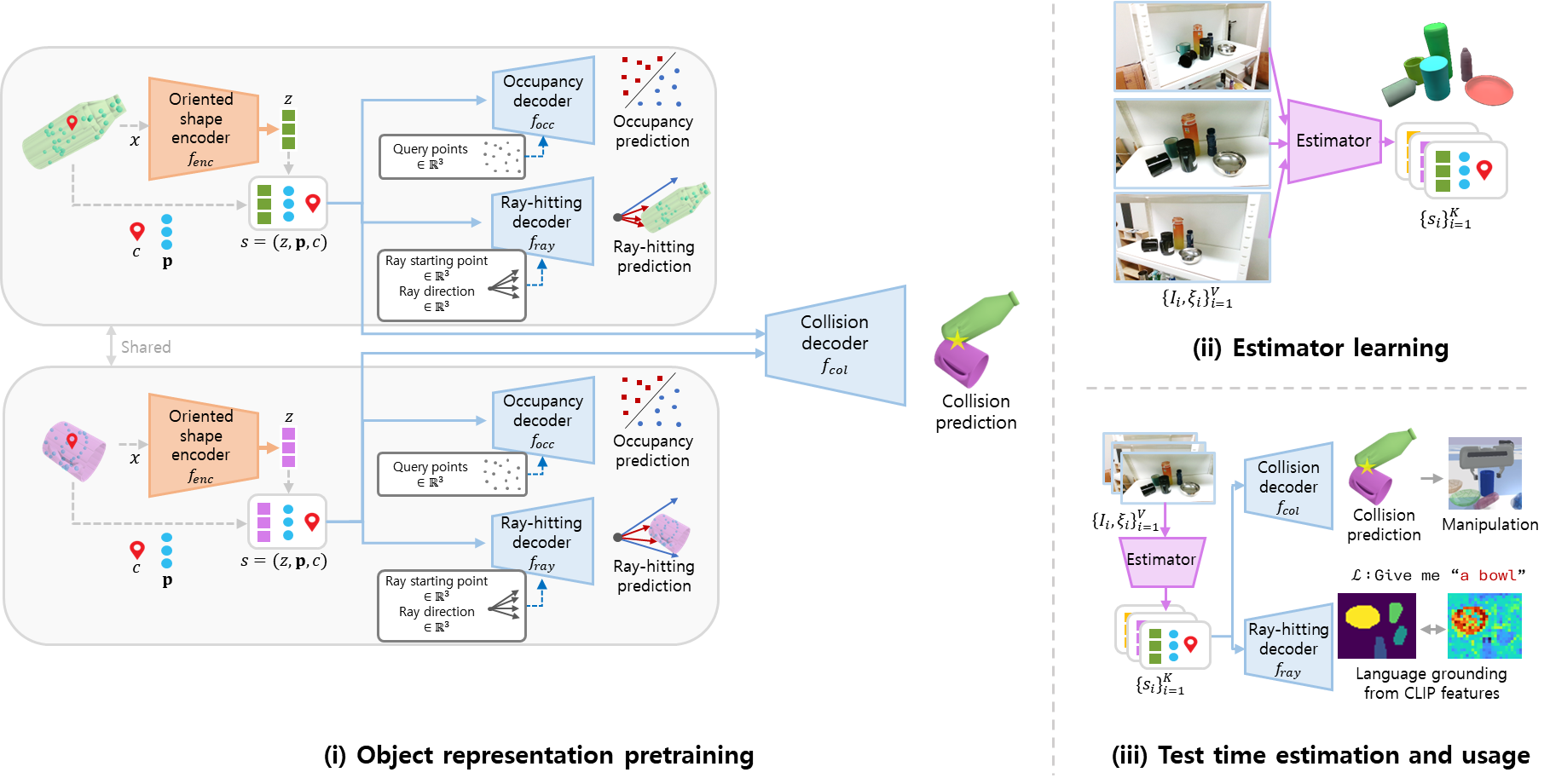}
    \caption{Three stages of our framework: (i) 
    An architecture for training the oriented shape embedding $z$. Given a mesh, we create its point cloud, $x$, and determine the location of the volumetric centroid, $c\in\mathbb{R}^3$, and $M$ geometrically representative points $\mathbf{p}\in\mathbb{R}^{M \times 3}$. The orange box is the encoder $f_{enc}$, which encodes $x$ into $z$, and the blue boxes are the decoders. The occupancy decoder $f_{occ}$ receives query points and $\mathbf{s}$ to output zero if the query points are inside the object. The ray-hitting decoder $f_{ray}$ receives $\mathbf{s}$, ray starting point, and direction and outputs a binary value indicating whether the ray hit the object. The collision decoder $f_{col}$ predicts the binary value indicating whether the pair of object states are in collision.
    (ii) In the estimator learning stage, we fix $f_{enc}$ and train the estimation module that predicts $\mathbf{s}$ from a set of $V$ images with intrinsic and extrinsic camera parameters $\{I_i,\xi_i\}_{i=1}^{V}$. (iii) In the test phase, we use estimated \(\mathbf{s}\) and the decoders, \(f_{col}\), and \(f_{ray}\), for collision checking and ray-testing, for motion planning and language grounding.}
    \label{fig:overview}
\end{figure*}


Our insight is that for most abstract \emph{language}-commanded object manipulation tasks, a numerical orientation of an object is unnecessary, provided that we can approximate the object's spatial occupancy and check collisions. Therefore, we propose an alternative representation in which we encode both the shape and orientation of an object in a neural representation $z$ and express its position with volumetric centroid $c \in \mathbb{R}^3$ and a set of $M$ representative geometric points $\mathbf{p} \in \mathbb{R}^{M\times3}$, and express an object state as $\mathbf{s}=(z,c,\mathbf{p})$. 

We pre-train the neural-oriented shape representation $z$ by extending OccNet~\cite{OccupancyNetworks}, a standard shape reconstruction algorithm for objects at a canonical orientation, to accommodate varying orientations. To do this, we use a SO(3)-equivariant network, which, unlike standard networks, guarantees equivariance
and allows the model to generalize to novel orientations even without data augmentation. In~\cite{son2024an}, we have proposed a SO(3)-equivariant network, FER-VN, that achieves state-of-the-art performance for shape prediction tasks, which we use in this work.


Checking collision with a neural representation, however, is expensive as it requires constructing a mesh from the representation. So, we propose to directly predict collisions from our representation by adding an extra decoder to our pre-training network that predicts the collision between the two objects A and B using their states $\mathbf{s}_A$ and $\mathbf{s}_B$ when training $z$. During motion planning or grasp selection, we use our collision decoder instead of an analytical collision checker~\cite{gjk}, eliminating the need for expensive mesh construction. Furthermore, thanks to our SO(3)-equivariant architecture, once we estimate object state representation $\mathbf{s}$, we can apply various SE(3) transforms to $\mathbf{s}$ to see if an object can fit inside a particular region, allowing efficient pick-and-place planning.

To ground the objects mentioned in the language command, we propose to use CLIP~\cite{clip}. In CLIP, an image encoder is trained such that an image embedding yields a high dot product with the corresponding text embedding. So, we can ground a text, say, ``a soda'', to an object state $\mathbf{s}$ by determining which pixels align with $\mathbf{s}$, aggregating the CLIP features at those pixels, and checking whether we have a large dot product with the soda text embedding. However, this incurs a high computational load because determining the alignment between pixels and object state $\mathbf{s}$ requires shooting a ray from each pixel, and checking if it hits $\mathbf{s}$ using the occupancy decoder. 



To resolve this, we add yet another decoder to our pre-training network that, given a ray and object state $\mathbf{s}$, predicts whether the given ray would hit $\mathbf{s}$. To create the CLIP feature for an object given an image, we evaluate the pixel-wise CLIP features and then run the ray-hitting decoder with a set of rays from each pixel to see if it hits the object's state $\mathbf{s}$. Altogether, our representation pre-training network involves occupancy, collision, and ray-hitting decoders, and we call the representation \textbf{ori}ented \textbf{\texttt{C}}ollision, \textbf{\texttt{O}}ccupancy, and \textbf{\texttt{R}}ay-hitting representatio\textbf{\texttt{N}} (\ourrep). The complete architecture is shown in Figure \ref{fig:overview} (left). 



%% file: intro_sec1-2.tex
\begin{figure*}[t]
    \centering
    \includegraphics[width=0.98\linewidth]{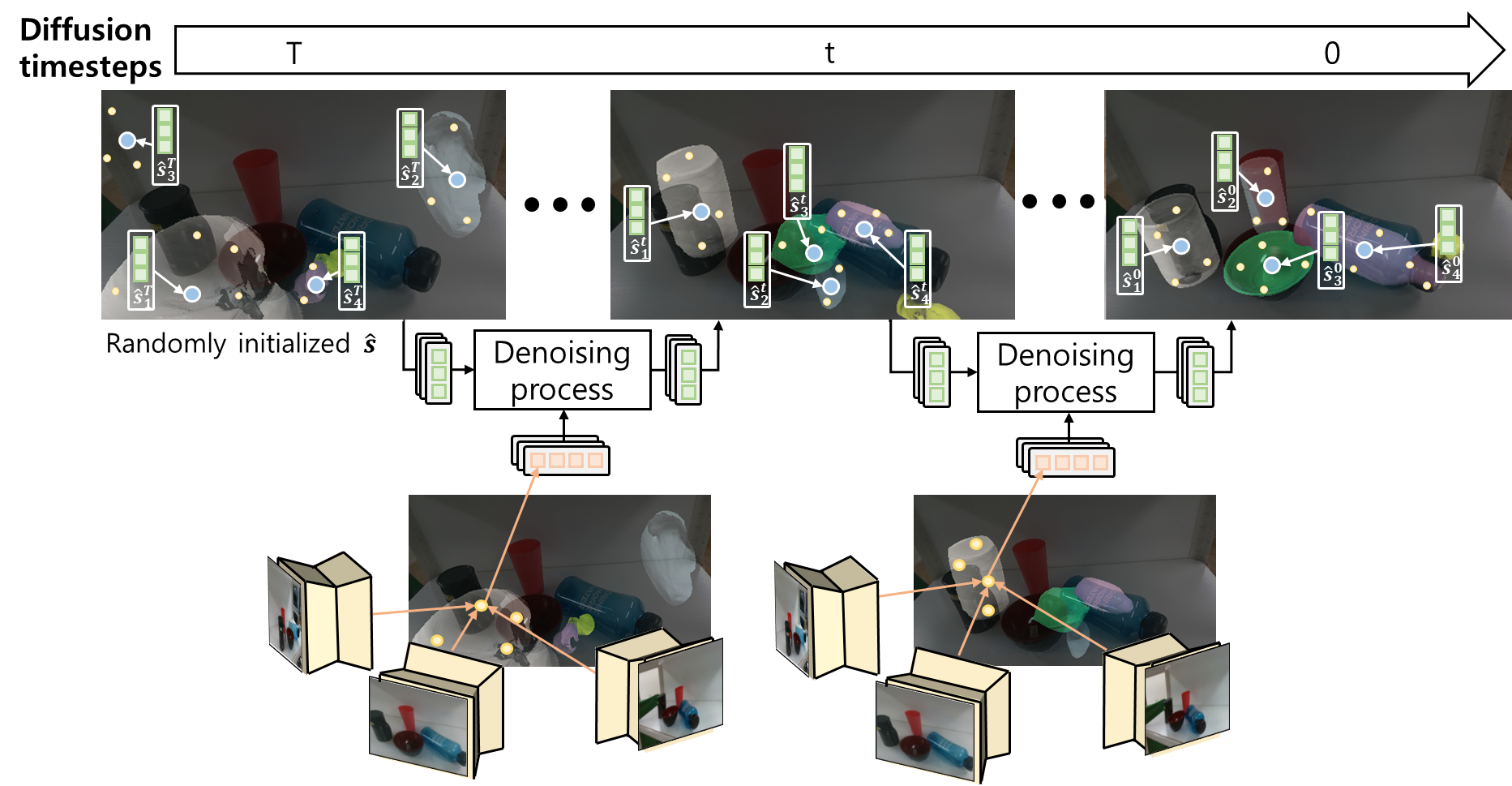}
    \caption{
    Our diffusion-model-based state estimator for \ourrep. The process begins with a random Gaussian initialization of $K$ number of object states at diffusion timestep $T$, denoted as $\{\hat{\mathbf{s}}_i^T\}_{i=1}^K$. Each $\mathbf{s}$ includes the center positions of objects and geometrically representative points, marked by blue circles and yellow circles respectively in the images. We iteratively refine $\hat{\mathbf{s}}$ by progressing through diffusion timesteps from \(T\) to 0, recursively applying the denoising process. 
    First, images are processed using a U-Net architecture~\cite{ronneberger2015unet} (yellow boxes) to extract pixel-wise image features. 
    At each denoising step, image features corresponding to the projected locations of the \(\mathbf{p}\) (illustrated as yellow circles), are extracted from the image plane, depicted as orange boxes. \(\{\hat{\mathbf{s}}_i^t\}_{i=1}^K\) alongside these image features are then jointly processed to refine object states to \(\{\hat{\mathbf{s}}_i^{t-1}\}_{i=1}^K\) for the following timestep \(t-1\).
    Upon the completion of \(T\) timesteps, the refined \(\{\hat{\mathbf{s}}_i^0\}_{i=1}^K\) are produced as the output.
    }
    \label{fig:entire_estimation}
\end{figure*}

The second core challenge is estimating object's \ourrep~from a sparse set of RGB images once we have pre-trained it. The naive way to do this is to apply 2D object detection on the input images and gather image features for each detected bounding box to estimate object states. However, this approach applies detection to each image independently and it is difficult to aggregate information across images obtained from multiple viewpoints~\cite{Wang-2019-117586}. 

Instead, we propose an iterative algorithm inspired by DETR3D~\cite{detr3d} that refines the object state estimation from a random initialization. At each iteration $t$, we use the current estimation $\mathbf{s}^t$ to project $\mathbf{p}^t$ from 3D to 2D pixel space, aggregate features from the corresponding pixels of each image, and then process the aggregated features to predict $\mathbf{s}^{t+1}$. This method allows integrating features from multiple images without having to solve the complex association problem, and integrates seamlessly into a diffusion model, which iteratively denoises a random noise into an estimand, that we use to characterize the uncertainty that stems from occlusion.  We call this 
\textbf{\texttt{D}}iffusion-based \textbf{\texttt{E}}stimator using aggregated \textbf{\texttt{F}}eatures for \ourrep\ (\entirename).
The estimation pipeline is shown in Figure \ref{fig:entire_estimation}.

The overview of \entirename~is given in Figure~\ref{fig:overview}. In both synthetic and real-world estimation, grasp planning, motion planning, and language-directed manipulation problems, we show that our method outperforms the state-of-the-art in terms of both speed and accuracy in estimation and planning.

%% file: related_works.tex
\section{Related Works}

 \subsection{3D object and scene representations}
There are several implicit-function-based representations for scenes and objects for manipulation. In compositional NeRF~\cite{driess2023learning} the authors show that you can plan simple object-pushing motions by representing each object with a NeRF and learning a dynamics model in the latent space. However, for scenes with a large number of objects, this approach would require a significant amount of time in both training and inference, since you need a NeRF for each object. In~\cite{shen2023distilled}, the authors propose an approach that learns a non-object-based scene representation with a single NeRF and uses a CLIP embedding~\cite{radford2021learning} to ground language command to perform pick-and-place operations. The crucial drawback of both of these approaches is that they require a large number of camera views of the scene (e.g. 50 images in~\cite{shen2023distilled}) and oracle demonstrations as it is difficult to apply traditional robot planning algorithms to NeRF-based representations.

There are extensions of NeRF that use a small number of camera views \cite{yu2021pixelnerf, chen2021mvsnerf, dai2023graspnerf}. These methods use CNN-based image feature extractors and NeRF decoders that are pre-trained with multiple scenes which enables scene reconstruction with only a few number of images. There also are several works that just use a single RGB-D image for object state estimation~\cite{irshad2022shapo, irshad2022centersnap}, in which you estimate the poses and shapes of multiple objects, where each object shape is represented with a Signed Distance Fields (SDFs). However, all of these approaches represent objects using occupancy-based implicit functions (e.g. NeRFs, SDFs) which are very expensive to use for manipulation. To detect a collision, you need to obtain the surface points of objects and then evaluate whether each point is inside another object using its implicit function~\cite{le2023differentiable}, which requires trading-off accuracy with computational speed. Alternatively, you can run a mesh creation algorithm, such as Marching Cubes~\cite{lorensen1998marching}, and then run an analytical collision checker but this also incurs a large computation time. Furthermore, RGB-D sensors are error-prone for shiny or transparent objects. 

There are several key-point-based object representations for manipulation. kPAM~\cite{kpam} represents objects with a set of keypoints annotated by humans, which are estimated from camera images and are then used for pick-and-place operations. 
This approach demonstrates that keypoints can generalize across unseen instances within the same category.
However, keypoints have limitations, such as inadequate representation of an object's full geometry and restricted generalization to novel object poses due to their sparse nature.
To address these limitations, the work in~\cite{simeonov2022neural} introduces SE(3) equivariant neural descriptive fields (NDFs).
These fields represent an object as a continuous function, mapping 3D coordinates to category-level spatial descriptors, offering a more detailed, volumetric representation compared to discrete keypoints. The use of SE(3)-equivariance of this approach also ensures that descriptors remain consistent relative to the object's transformation, effectively generalizing over all possible translations and rotations.
This approach, however, requires expert demonstrations, which are expensive to obtain.
Additionally, it requires depth sensors to get the point cloud, limiting its use ofr shiny or transparent objects.




CabiNet~\cite{murali2023cabinet} and Danielczuk et al.~\cite{danielczuk2021object} train a neural contact detector that detects contact between a scene point cloud and an object of interest. Similar to us, the neural contact detector is integrated with conventional motion planning algorithms to compute collision-free paths  Our representation is closest to \cite{cho2024corn}, where the authors propose a contact-based object representation based on depth data. The representation is used to learn a contact-rich non-prehensile manipulation policy that generalizes across different object shapes. Our representation is much inspired by this work, but we extend this to arbitrary object pairs, not just between robot grippers and objects, and show that we can estimate the 3D location and object shape using RGB images.

In \cite{wang2023d3fields}, authors propose an approach where they first obtain object point clouds through instance segmentation, compute pixel-wise 2D semantic features using the pre-trained Grounding-DINO model~\cite{liu2023grounding}, and project them onto the surface points of each 3D object point cloud and use this for manipulation. Since such object representation is defined with the explicit object point cloud, estimating interactions between objects, such as collisions, necessitates additional computation for mesh generation or training an independent neural dynamics model.

\begin{figure}[t]
    \centering
    \includegraphics[width=0.95\linewidth]{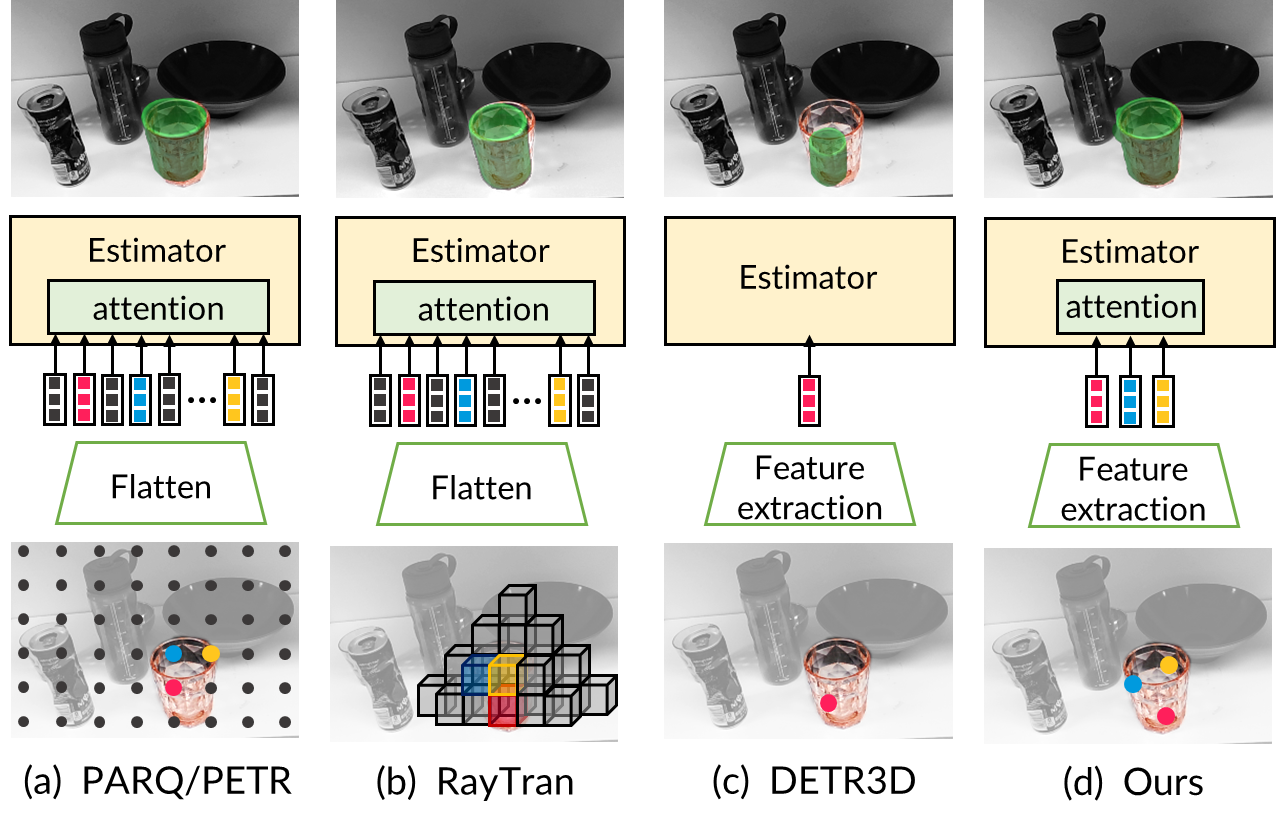}
    \caption{
    Comparison of different image feature extraction methods from different estimation methods: 
    (a) PARQ and PETR utilize image features from all pixels
    and use an attention mechanism on the extracted features.
    (b) RayTran uses voxels to extract image features.
    It processes images with a 2D CNN, defines a 3D voxel grid on a scene, projects each grid cell to 2D image planes, extracts image features at those locations, and utilizes attention mechanism on these features. The voxel grid is visualized in the figure. 
    (c) DETR3D extracts image features from a single pixel.
    It iteratively refines object location predictions and uses these predictions to extract relevant image features. The red point indicates a predicted object location during an intermediate iteration, 
    which is then used to get the image feature at that 3D point using projection.
    (d) Contrary to using a single point for feature extraction, our method gathers image features from a set of geometric representative points $\mathbf{p}$, illustrated by three points.
    }
    \label{fig:network_comparison}
\end{figure}

\subsection{Object localization and shape estimation algorithms}
Recently, several works have proposed object detection algorithms that estimate 3D object locations using a small number of RGB cameras with known poses~\cite{raytran,xie2023pixel,liu2022petr,detr3d}. Much like our \entirename, these methods leverage camera poses to extract image features and use them to predict object locations iteratively, but they differ in how they extract the features from images as shown in Figure \ref{fig:network_comparison}. 

PETR~\cite{liu2022petr} and PARQ~\cite{xie2023pixel} utilize pixel-wise image features across the entire scene (Figure \ref{fig:network_comparison}-(a)), but this is inefficient because it processes irrelevant features too, such as those from the empty space, shown in black dots in Figure \ref{fig:network_comparison}-(a). Alternatively, we can use features from pixels that correspond to cells of a pre-defined voxel grid~\cite{raytran} (Figure\ref{fig:network_comparison}-(b)) but this method must use high voxel grid resolution to achieve high accuracy and similarly struggles with the inefficiency of processing irrelevant grid cells. Our method is most similar to DETR3D~\cite{detr3d}, which uses the center point for each object to get the corresponding image feature, making it more computationally efficient than the approaches mentioned ealrier. One important distinction, however, between our method and DETR3D is the number of pixel-wise features: we use pixels that correspond to \emph{multiple} points in $\mathbf{p}^t$, while DETR3D uses the pixel that corresponds to a \emph{single} point, which proved insufficient for capturing the full complexity of accurate object geometry~\cite{liu2022petr}. The comparison is shown in Figure~\ref{fig:network_comparison}-(c) and (d). All these methods are primarily proposed for object localization and do not infer shapes. More importantly, these methods cannot characterize the uncertainty in shape and location estimation caused by occlusions. 

There are several Oject Simultaneous Localization and Mapping (Object SLAM) algorithms for estimating object shapes and locations. NodeSLAM \cite{sucar2020nodeslam} and NeuSE \cite{fu2023neuse} use an occupancy encoder to encode object shape into a latent embedding, and optimize over the embedding and object and camera poses by minimizing the discrepancy from depth observations. These methods address a different problem than ours in that camera locations are also inferred from a sequence of observations. While we can apply these methods to our problem, their use of implicit shape functions and depth sensors renders them expensive to use and error-prone to shiny or transparent objects. 

Recently, several methods have been proposed for shape prediction from a single image~\cite{liu2024one, hong2024lrm}. One-2-3-45~\cite{liu2024one} has two steps: 
(1) generating multi-view images from a single image and a relative camera pose using a view-conditioned 2D diffusion model, Zero123, and 
(2) training a NeRF with these generated multi-view images.
Applying the Marching Cubes algorithm to the SDF field predicted by the NeRF, it produces a high-quality reconstructed surface. However, the process of generating multiple images and training a NeRF is time-consuming, making real-time execution difficult. On the other hand, Large Reconstruction Model (LRM)~\cite{hong2024lrm} extracts tri-plane representations from a single image and camera pose, and uses them to pre-train a neural shape representation with a NeRF-based decoder and image reconstruction loss. This approach is faster than One-2-3-45 but it still requires approximately five seconds for each object shape prediction~\cite{lorensen1998marching}, rendering them impractical for real-time multi-object shape estimation.

\subsection{SO(3) equivariant neural networks}
Several practical algorithms have recently been proposed for SO(3)-equivariant neural networks~\citep{deng2021vector, puny2021frame, kaba2023equivariance,qi2017pointnet, wang2019dynamic}. Frame averaging introduced in~\cite{puny2021frame} implements equivariance by averaging over a group using symmetry, but its use of Principal Component Analysis (PCA) makes it vulnerable to noise and missing data, which are common in 3D data. Kaba et al.~\cite{kaba2023equivariance} propose an approach that uses an additional network that aligns rotated inputs to a canonical pose and then applies standard point processing networks. The limitation here is that a small error in the alignment network could have a large negative impact on the overall performance.  Vector Neurons~\cite{deng2021vector} makes minor modifications to existing neural network operations, such as pooling operations and activation functions, to implement equivariance and does not suffer from the aforementioned problems. One limitation, however, is that it is limited to three-dimensional feature spaces, which makes it difficult to encode high-frequency 3D shape information. Our recent work~\cite{son2024an} addresses this issue by proposing a provably equivariant high-dimensional feature representation, FER, that can capture a variety of different frequencies, and proposes FER-VN. We use this to pre-train our representation~\ourrep.

\subsection{Generative models for characterizing uncertainty in shape prediction}
Estimating 3D shapes from a sparse number of views is challenging due to the inherent uncertainty in object shapes and poses caused by occlusion. To address this issue, several approaches utilize generative models such as a Variational Auto-Encoder (VAE) or diffusion model to express the uncertainty. Simstack~\cite{landgraf2021simstack} employs a depth-conditioned VAE, trained on a dataset of stacked objects, to infer object shapes expressed using truncated signed distance function (TSDF). The decoder of the VAE estimates the complete object shape from partial TSDF. In~\cite{wu2020multimodal}, the authors use a conditional Generative Adversarial Network (cGAN) to express the multimodality present in completing shapes from occluded inputs, and learn a lower-dimensional manifold of shapes using an auto-encoder. Given a partial point cloud, the generator of cGAN is then trained to output a sample from this lower-dimensional manifold. Diffusion models~\cite{firstdiffusionmodel} have recently emerged as an alternative for expressing shape reconstruction and generation, due to their stability in training and the ability to characterize multimodality, and has shown success in 3D shape generation tasks~\cite{zhou20213d, cheng2023sdfusion}.




%% file: method.tex
\section{Problem formulation}

\newcommand{\image}{I}
\newcommand{\camparam}{\xi}
\newcommand{\objectrepresentation}{\mathbf{s}}
\newcommand{\cliprep}{\mathbf{e}}
\newcommand{\numcam}{V}
\newcommand{\numobjs}{K}
\newcommand{\langcmd}{\mathcal{L}}
\newcommand{\orientedshape}{z}
\newcommand{\representativepoints}{\mathbf{p}}
\newcommand{\centerpoint}{c}
\newcommand{\zdim}{u}
\newcommand{\pdim}{m}
\newcommand{\evalscene}{\text{IoU}}
\newcommand{\evalsceneset}{\{\text{IoU}\}}

We assume that we are given a $V$ number of RGB images, $\{\image_i\}_{i=1}^\numcam$, intrinsic and extrinsic camera parameters $\{\camparam_i\}_{i=1}^\numcam$, and a language command $\langcmd$. Our objective is to develop an object state representation learning and estimation algorithms for language-directed manipulation planning. This consists of the following three sub-problems:
\begin{itemize}
\item \textbf{Object representation learning (Section \ref{sec:construct_geometric_objrep})}: how to train our object state representation, \ourrep, that facilitates efficient collision checking and language grounding.

\item \textbf{Estimator learning (Section \ref{sec:generative_model})}: how to train a diffusion-model-based estimator which takes as inputs $\{\image_i,\camparam_i\}_{i=1}^{\numcam}$ and outputs a distribution over $\{\hat{\objectrepresentation}_i\}_{i=1}^{K}$ an estimation of ground truth object state $\{\objectrepresentation_i\}_{i=1}^{K}$.

\item \textbf{Language grounding using CLIP (Section \ref{sec:clip_overlay})}: given $\{\image_i,\camparam_i\}_{i=1}^{\numcam}$ and $\{\hat{\objectrepresentation}_i\}_{i=1}^{\numobjs}$, how to ground object text to $\{\hat{\objectrepresentation}_i\}_{i=1}^{\numobjs}$.
\end{itemize}
All training is done in simulation using synthetic data, and zero-shot transferred to the real-world. Figure~\ref{fig:overview} gives an overview of our framework~\entirename.

\section{Pre-training~\ourrep} \label{sec:construct_geometric_objrep}
Our object state representation, \ourrep, denoted $\objectrepresentation$, consists of three quantities: oriented shape embedding, $\orientedshape\in \mathbb{R}^\zdim$, volumetric center, $\centerpoint \in \mathbb{R}^3$, and geometrically representative points $\representativepoints \in \mathbb{R}^{\pdim \times 3}$. We only train $\orientedshape$, and $c$ and $\representativepoints$ are analytically computed.
To train $\orientedshape$, we prepare a dataset of the form
\begin{align*}\{(x_u, c_u, \representativepoints_u, q_{occ,u}, q_{ray,u}, d_u)_{u\in\{A,B\}}, \\ (y_{col}, y_{occ,A}, y_{occ,B}, y_{ray,A}, y_{ray,B})\} \,
\end{align*}
where $u\in\{A,B\}$ is an object index, $x_u$ is a surface point cloud, $q_{occ,u} \in \mathbb{R}^3$ is the query point for the occupancy decoder, $q_{ray,u} \in \mathbb{R}^3$ and $d_u \in \mathbb{R}^3$ are the origin and the direction of the ray respectively, $y_{col}$ is the collision label between objects $A$ and $B$,  $y_{occ,u} \in \{0,1\}$ is the occupancy label at $q_{occ,u}$, and $y_{ray,u}$ is the ray-test label for $(q_{ray,u}, d_u)$.

To prepare our dataset, we randomly select a pair of object meshes from a 3D shape dataset~\cite{Wang_2019_CVPR_NOCS}, randomly scale them, and place them in random poses within a bounded 6D space. We then sample its surface point cloud $x$, calculate its volumetric centroid $c$, and apply convex decomposition to get the center of each convex hull for geometrically representative points $\representativepoints$. We use GJK~\cite{gjk} to get collision label $y_{col}$. For $q_{occ}$, we sample surface points and add Gaussian noise, and for $y_{occ}$, we evaluate whether $q_{occ}$ is inside of the object. 
To simulate rays for training $f_{ray}$, we randomly select a direction $d$ uniformly from the unit sphere in $\mathbb{R}^3$. We then sample a surface point, add small Gaussian noise, and calculate the starting point of the ray $q_{ray}$ by subtracting $d$ scaled by a predefined large distance from the perturbed surface point. This process effectively positions $q_{ray}$ to ensure it originates sufficiently far from the object while maintaining its directional integrity towards the object. Given $d, q_{ray}$ and the object mesh, we evaluate a ray-hitting result by analytical mesh-ray hitting test~\cite{roth1982ray} to get $y_{ray}$.


Figure \ref{fig:overview} (left) shows our architecture. The oriented shape encoder $f_{enc}$ outputs $\orientedshape$, and the collision decoder $f_{col}$ takes in $\objectrepresentation=(\orientedshape,\centerpoint,\representativepoints)$ for each object, and outputs a binary value indicating the presence of collision. The ray-hitting decoder ($f_{ray}$) processes $\objectrepresentation$ for an individual object alongside a ray defined by ray starting point $q_{ray}$ and ray direction $d$, and predicts whether the ray hits the object. Similarly, the occupancy decoder ($f_{occ}$) takes $\objectrepresentation$ and a query point $q_{occ}$ and predicts whether the point lies inside the object. 

We have the following loss function for training our representation
\begin{align*}
    L &= \lambda_{\text{col}}\sum_{i}L_{\text{col}}(x_A^{(i)},x_B^{(i)},\centerpoint_A^{(i)}, \centerpoint_B^{(i)},\representativepoints_A^{(i)},\representativepoints_B^{(i)}   ,y^{(i)}_{\text{col}}) \nonumber \\
    &+ \lambda_{\text{occ}}\sum_{u \in \{A,B\}}\sum_{j} L_{\text{occ}}(x_u^{(k)},c_u^{(k)},\representativepoints_u^{(k)},q^{(j)}_{occ,u},y^{(j)}_{\text{occ},u}) \nonumber \\
    &+ \lambda_{\text{ray}} \sum_{j \in \{A,B\}}\sum_{k}L_{\text{ray}}(x_j^{(k)},c_j^{(k)},\representativepoints_j^{(k)},q^{(k)}_{ray,j},d^{(k)}_j,y^{(k)}_{\text{ray},j}) 
\end{align*}
where $L_{col}$, $L_{occ}$, $L_{ray}$ are losses for collision detection, occupancy decoder, and ray-hitting decoder respectively, each of which are binary cross entropy function, $\lambda_{col}, \lambda_{occ}, \lambda_{ray}\in\mathbb{R}$ are coefficients for each loss.
$x^{(i)}$ is an object surface point cloud in the $i^{th}$ data point placed at a random location in a 3D environment.


\begin{figure}[t]
    \centering
    \includegraphics[width=0.99\linewidth]{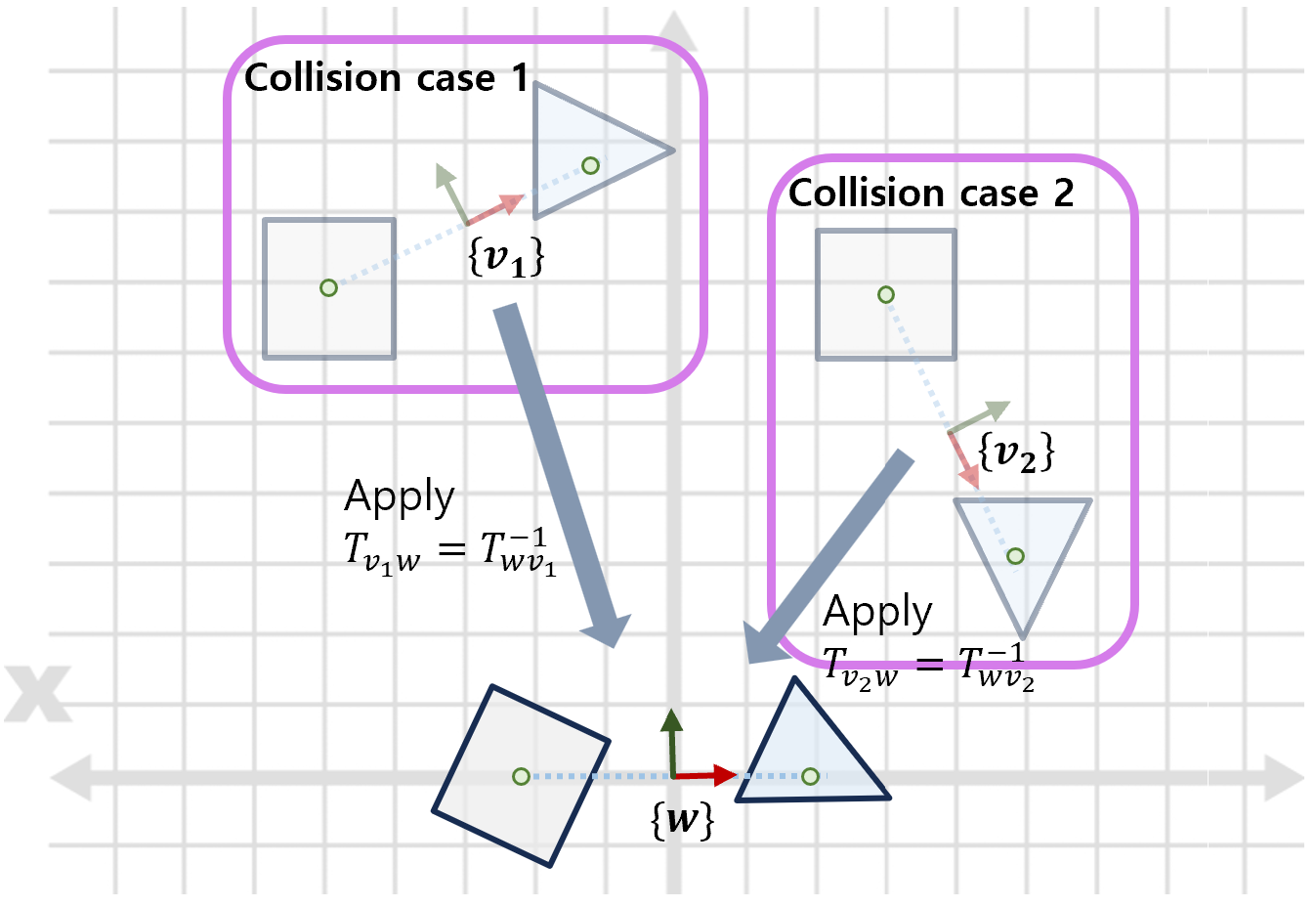}
    \caption{Illustration of pre-processing for achieving invariance. Case 1 and 2 have the same relative transform between two objects, but their global transforms are different. If we treat the two objects as a single composite rigid body, we can assign frames $\{v_1\}$ and $\{v_2\}$ whose origin is at the mid-point of the centers of two objects, and whose direction is determined by the line intersecting the centers. We then apply $T_{v_1w}$ and $T_{v_2w}$ to these frames so that they are at the origin of the world frame $\{w\}$, with their orientation aligned with that of $\{w\}$ as shown in the bottom. This preprocessing step ensures consistent input irrespective of objects' global poses.
}
    \label{fig:con_invariance}
\end{figure}

One important factor in all three decoders is that if the relative transform between the two objects (in the case of collision decoder) or between the object and query vector (in the case of ray and occupancy decoders) is the same, then the decoded output should stay the same even if their global transform changes. Unfortunately, standard neural networks cannot do this without excessive data augmentation.


Therefore, we use an SO(3)-equivariant network, FER-VN~\cite{son2024an} for our encoder, which allows us to apply transformations to $\orientedshape$ and make our decoders invariant as long as objects have the same relative locations. For $f_{occ}$ and $f_{ray}$, we use the FER-VN-OccNet decoder to achieve invariance as done in \cite{deng2021vector, son2024an}.
Specifically, we use two independent SO(3)-equivariant networks, \(g_a\) and \(g_b\), to process the shape embedding \(z\) and 3D query points \(q\), respectively, ensuring rotational equivariance such that \(g_a(Rz) = Rg_a(z)\) and \(g_b(Rq) = Rg_b(q)\) for any rotation \(R\). The networks' outputs are combined through an inner product, \(g_a(z) \cdot g_b(q)\), to maintain rotational invariance, with a subsequent Multilayer Perceptron (MLP) generating the final occupancy or ray-hitting predictions.
However, for the collision decoder, the input is a pair of object representations, and using the FER-VN-OccNet decoder is not trivial. Therefore, we design a pre-processing step inside $f_{col}$ that makes the collision decoder invariant to the same relative location between a pair of objects. The procedure is described in Figure \ref{fig:con_invariance}. 

\section{Learning and using the estimator}
\label{sec:generative_model}

\subsection{Estimator learning}
\begin{figure}[t]
    \centering
    \includegraphics[width=1\linewidth]{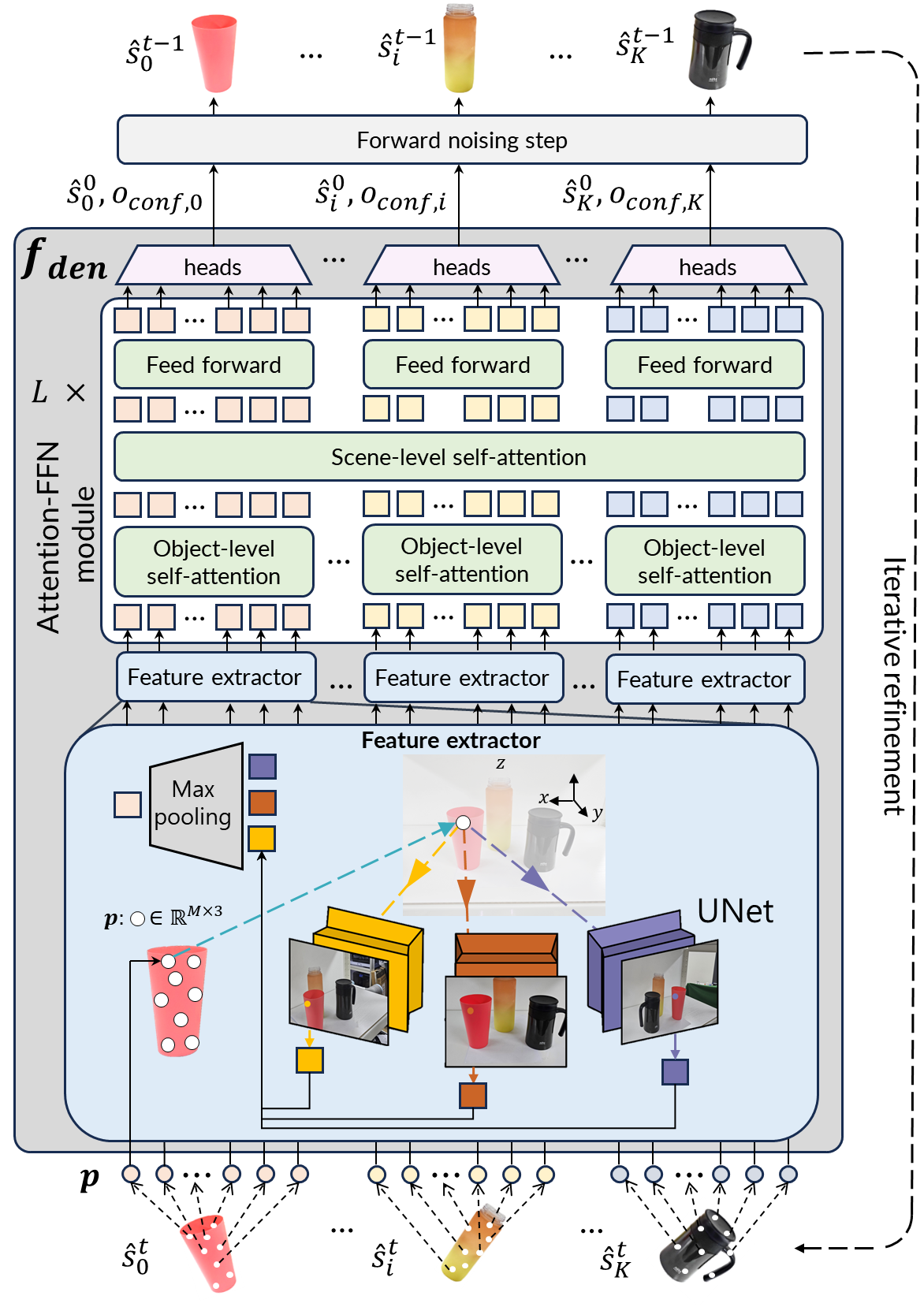}
    \caption{Architecture and computational structure for our diffusion-based estimator $f_{den}$. $f_{den}$ takes in the estimate of the previous diffusion timestep, $\hat{\objectrepresentation}^{t}$, and images to outputs a refined $\hat{\objectrepresentation}^{t-1}$. To do this, we utilize $\representativepoints \in \mathbb{R}^{M\times3}$, shown with a small circle at the bottom, to gather image features corresponding to each object by a feature extractor (shown as blue boxes). The feature extractor first generates pixel-wise image features by processing images with U-Net~\cite{ronneberger2015unet} and for each point in $\representativepoints$, we aggregate features by projecting $\representativepoints$ into the image plane and performing max-pooling across multiple views. So, we get one image feature vector for each point in $\representativepoints$ across images, and end up with $M$ number of feature vectors for each object. The figure illustrates this for a particular point from $\representativepoints$ for the pink cup. These features are then processed by the attention feed-forward network (attention-FFN) module (the upper white box) consisting of object-level attention, which processes features for each object independently, scene-level self-attention, which processes features across objects, and FFN. We repeat the attention-FFN module $L$ times, followed by heads to output $\hat{\objectrepresentation}^{0}$ and $o_{conf}$, the confidence of the prediction.  $\hat{\objectrepresentation}^{0}$ is processed by forward noising step to get $\hat{\objectrepresentation}^{t-1}$, which is recurrently fed into $f_{den}$ again.}
     \label{fig:network_structure}
\end{figure}

Our goal is to learn an estimator that predicts $\objectrepresentation$ for each object in a scene from multiple RGB images with known poses.  To characterize the uncertainty caused by partial observability, we use a diffusion model and implement a denoising function $f_{den}$ as shown in Figure \ref{fig:network_structure}.

The function takes a set of $N$ (the maximum number of the objects on the scene) object representations $\{\hat{\objectrepresentation}^{t}_i\}_{i=1}^{N}$ at diffusion timestep $t$ and images and camera parameters $\{I_i,\xi_i\}_{i=1}^V$, and iteratively refine $\hat{\objectrepresentation}^t$ from $t=T$ to 0 to output $\{\hat{\objectrepresentation}^{0}_i\}_{i=1}^{N}$ where $T$ is the number of diffusion time steps. We design $f_{den}$ to predict denoised object state representation $\{\hat{\objectrepresentation}^{0}_i\}_{i=1}^N$ directly instead of noise following previous works \cite{karras2022elucidating, chen2023diffusiondet, Ramesh2022HierarchicalTI}. $\hat{\objectrepresentation}^{t-1}$ can be obtained by adding noise to $\hat{\objectrepresentation}^{0}$, whose scale is determined by $t-1$, via a forward noising step.





To prepare the training dataset, we create a scene in simulation (e.g. Figure \ref{fig:synthetic_data}) and create a data point of the form $\{(I_i,\xi_i)^V_{i=1},(\objectrepresentation_i)^K_{i=1}\}$, where $K$ is the true number of objects in the scene and $\objectrepresentation$ is the target object state representation, obtained by feeding the the ground-truth object point cloud to $f_{enc}$ and analytically computing $c$ and $\representativepoints$.  We denote target object state representation at diffusion time step $t$ as $\objectrepresentation^t$, so $\objectrepresentation^0$ is target object representation and $\objectrepresentation^T$ is a Gaussian noise.
Given $\objectrepresentation^0$, we first sample diffusion time step $t$ from a uniform distribution, and add a Gaussian noise, whose scale is determined by $t$, to $\objectrepresentation^0$ to get $\objectrepresentation^t$ using a forward noising step. We also make $f_{den}$ to output $o_{conf,i}$, the confidence of estimation for object $i^{th}$. Then, we minimize the following loss
\begin{align}
\label{eq:dif_loss}
\sum_{i,j\in \text{bipartite}(\{\hat{\objectrepresentation}^0_i\}^N_{i=1}, \{\mathbf{s}^0_j\}_{j=1}^K)}&{dist(\hat{\mathbf{s}}^0_i, \mathbf{s}^0_j) + BCE(o_{conf,i}, 1)} \notag\\
+\sum_{i\in \{i|\text{not matched}\}}&{BCE(o_{conf,i}, 0)}
\end{align}

where \textit{bipartite} is a function that determines bipartite matching between two sets based on the distance function $dist$, $BCE$ is binary cross-entropy, and $\{i|\text{not matched}\}$ is a set of indices which are not matched by the bipartite function.
We define the distance between two object states as
\begin{equation}
\label{eq:s_distance}
dist(\objectrepresentation, \objectrepresentation')=\alpha_p CD(p,p') + \alpha_c|c-c'|_F^2 + \alpha_z|z-z'|_F^2.
\end{equation}
where $\objectrepresentation=(z,c,\mathbf{p}), \objectrepresentation'=(z',c',\mathbf{p}')$, $|\cdot|_F$ is Frobenius norm, $CD$ is Chamfer distance, and $\alpha_{p},\alpha_{c},\alpha_{z} \in \mathbb{R}$ are hyperparameters.

As described in \cite{ji2023ddp}, our model is also susceptible to sampling drift where samples generated by $f_{den}$, denoted as $\hat{\mathbf{s}}^t$, may deviate from the target created during the training, $\mathbf{s}^t$, and this deviation can lead the denoising neural network to encounter $\objectrepresentation^t$ outside its training distribution. To mitigate this issue, we use a strategy that utilizes the reverse denoising process rather than the forward noising process. The major difference from standard diffusion training~\cite{ho2020denoising} is that we feed $\hat{\objectrepresentation}^t$ to $f_{den}$ instead of $\objectrepresentation^t$ obtained by adding noise to $\objectrepresentation^0$.  Specifically, we select $h$ timesteps uniformly at random and then order them in descending sequence to create the series $(t_1, t_2, ..., t_h)$. Starting at $t_1$, we apply the forward noising step to $\objectrepresentation^{0}$ using $t_1$ to get $\objectrepresentation^{t_1}$, and predict $\hat{\objectrepresentation}^{t_2}$ using $\objectrepresentation^{t_1}$ and $f_{den}$. Again, $\hat{\objectrepresentation}^{t_2}$ is fed into denoising network $f_{den}$ to output $\hat{\objectrepresentation}^{t_3}$. This recursive approach allows us to compile a set of states, ${(\hat{\mathbf{s}}^{t_2},...,\hat{\mathbf{s}}^{t_h})}$, for training. We feed these into $f_{den}$ to predict the denoised state, $\hat{\mathbf{s}}^0$, and compute the loss against the target data points, $\objectrepresentation^0$, as shown in equation \eqref{eq:dif_loss}. 
This method helps mitigate the deviation from $s^0$ during inference by continually guiding the potentially deviated state $\hat{s}^t$ back towards $s^0$.
Note that the procedure becomes standard diffusion training when $h=1$. 


\begin{algorithm}[t]
\caption{DiffusionInference($\{I_i, \xi_i\}_{i=1}^V, N, T, t_{step}, \epsilon$)}
\begin{algorithmic}[1]
\label{alg:diffusion_estimator}

\STATE $\mathcal{S}_{T} \gets \text{StandardGaussian}(0,1)$
\FOR{$t = T$ \TO $0$ \textbf{with $t_{step}$ interval}}
    \STATE $\mathcal{S}_0, \mathcal{O}_{\text{conf}} \gets f_{\text{den}}(\mathcal{S}_{t}, \{I_j, \xi_j\}_{j=1}^V, t)$
    \STATE $\mathcal{S}_0 \gets \text{ReplacewithStandardGaussian}(\epsilon, \mathcal{O}_{conf})$
    \STATE $\mathcal{S}_{t-t_{step}} \gets \text{ForwardNoising}(\mathcal{S}_0, \mathcal{S}_{t}, t, t-t_{step})$
\ENDFOR
\\
// Filter estimates with low confidence
\FOR{$n = 1$ \TO $N$}
    \IF{$\mathcal{O}_{\text{conf}}[n] \geq \epsilon$}
        \STATE $\mathcal{S}.\text{insert}(\mathcal{S}_{0}[n])$
    \ENDIF
\ENDFOR
\RETURN $\mathcal{S}$
\end{algorithmic}
\end{algorithm}

\subsection{Estimating the object states}
\label{sec:diffusion_estimation}
Algorithm~\ref{alg:diffusion_estimator} shows how to estimate states using our diffusion model. It
takes in images and camera parameters $\{I_i,\xi_i\}_{i=1}^V$, the maximum number of objects $N$, number of diffusion timesteps $T$, diffusion time step interval for DDIM~\cite{song2020denoising} sampling $t_{step}$, and confidence score threshold $\epsilon$ as inputs. The algorithm begins by initializing object states at diffusion timestep $T$, $\mathcal{S}_{T} := \{\hat{\mathbf{s}}^{T}_i\}_{i=1}^N$, with standard Gaussian noise (L1). After the initialization, lines 2-6 alternate between denoising and noising steps: the denoising step (L3) predicts \(\mathcal{S}_0\) from \(\mathcal{S}_t\) using \( f_{\text{den}} \), and the noising step adds noise to \(\mathcal{S}_0\) (L5), producing \(\mathcal{S}_{t-t_{\text{step}}}\) for the next iteration, as done in DDIM sampling process~\cite{song2020denoising}. In each time step, we filter $\objectrepresentation$ that has a low confidence score by replacing low-confidence representations with random Gaussian noise (L4). After this process, object representations with confidence values lower than \(\epsilon\) are removed (lines 7-11), resulting in the final set of object representations \(\mathcal{S}\), where $\mathcal{O}_{conf}[n]$ denotes the $n$th element of $\mathcal{O}_{conf}$.

\begin{algorithm}[t]
\caption{Refine($\{I_i, \xi_i\}_{i=1}^V, N, T, $ \\ $t_{step}, B, \epsilon, W, H$)}
\begin{algorithmic}[1]
\label{alg:estimation}

\STATE Initialize a set of $\mathcal{S}$: $\mathcal{C} \gets \emptyset$
\STATE Initialize a set of evaluations: $\{\text{IoU}\} \gets \emptyset$
\STATE $\{\mathcal{M}_i\}_{i=1}^V \leftarrow f_{\text{seg}}(\{I_i\}_{i=1}^V)$
\FOR{$b = 1$ \TO $B$ \textbf{in parallel}}
    \STATE $\mathcal{S} \leftarrow \text{DiffusionInference}(\{I_i,\xi_i\}_{i=1}^V, N, T, t_{step}, \epsilon)$
    \FOR{$w = 1$ \TO $W$}
        \STATE $\mathcal{S} \leftarrow \mathcal{S} + \lambda \nabla_{\mathcal{S}} \text{Evaluate}(\{\mathcal{M}_i\}_{i=1}^V, \mathcal{S})$
    \ENDFOR
    \STATE Store refined $\mathcal{S}$: $\mathcal{C}[b] \leftarrow \mathcal{S}$
    \STATE $\evalsceneset[b] \gets \text{Evaluate}(\{\mathcal{M}_i\}_{i=1}^V, \mathcal{S})$
\ENDFOR
\STATE Select the top-H: 
$\{\mathcal{S}_{i}\}_{i=1}^H \gets \text{TopH}(\mathcal{C}, \{IoU\},H)$
\RETURN $\{\mathcal{S}_{i}\}_{i=1}^H$
\end{algorithmic}
\end{algorithm}

\begin{algorithm}[t]
\caption{Evaluate($\{\mathcal{M}_i\}_{i=1}^V, \mathcal{S}$)}
\begin{algorithmic}[1]
\label{alg:evaluatescene}
\STATE Initialize: $\evalsceneset \gets \emptyset$ \\
 // Iterate over views
\FOR{$i = 1$ \TO $V$ \textbf{in parallel}} 
    \STATE $\hat{\mathcal{M}}_i \leftarrow \text{PixelwiseRayHitting}(\mathcal{S}, \xi_i)$
    \STATE $\evalscene_i \leftarrow \text{CalculateIoU}(\mathcal{M}_i, \hat{\mathcal{M}}_i)$
    \STATE $\evalsceneset\text{.insert}(\evalscene_i)$
\ENDFOR
\STATE Compute average IoU score: $\evalscene \leftarrow \frac{1}{V} \sum_{i=1}^V \evalscene_i$
\RETURN $\evalscene$
\end{algorithmic}
\end{algorithm}

\begin{figure*}[t]
    \centering
    \includegraphics[width=0.98\linewidth]{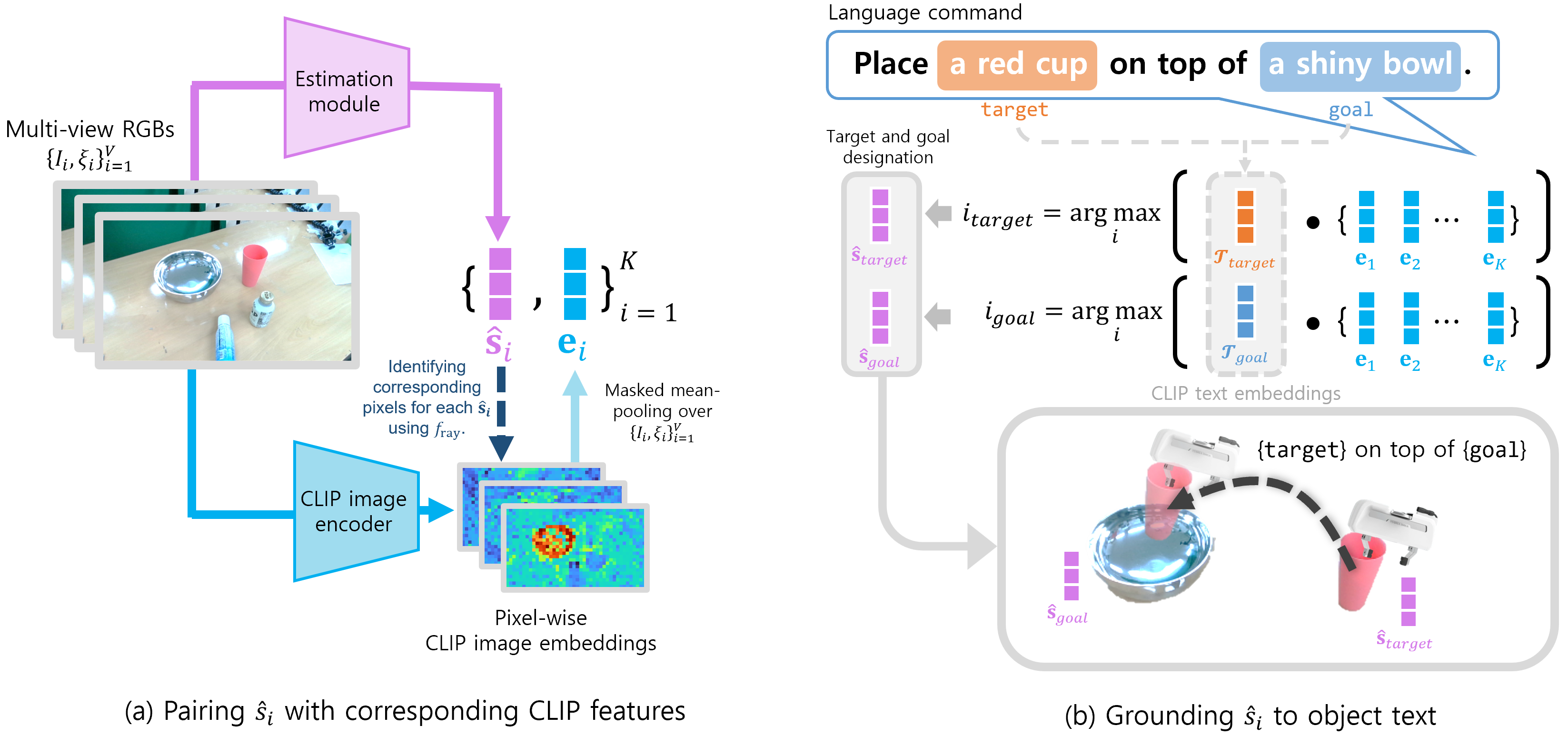}
    \caption{
    Overview of language-commanded pick-and-place pipeline. 
    \textbf{(a) Pairing $\objectrepresentation_i$ with corresponding CLIP features}: The estimation module estimates $\{\hat{\mathbf{s}}_i\}_{i=1}^K$ (pink arrow), and we process \(\{I_i\}_{i=1}^V\) using a CLIP encoder to obtain pixel-wise CLIP image embeddings (blue arrow). Next, we use \(f_{\text{ray}}\) to identify which pixels are occupied by each \(\hat{\mathbf{s}}_i\). For each object represented by \(\hat{\mathbf{s}}_i\), we collect the CLIP features from the occupied pixels and mean-pool them to derive the language-aligned representation \(\mathbf{e}_i\) for \(\hat{\mathbf{s}}_i\). 
    \textbf{(b) Grounding $\objectrepresentation_i$ to an object text}: Given a language command in a specified template, we first extract the target and goal object text, such as ``a red cup" and ``a shiny bowl," respectively. We then apply the CLIP text encoder to both the target and goal text to obtain the CLIP text embeddings $\mathcal{T}_{\text{target}}$ and $\mathcal{T}_{\text{goal}}$.
    For target and goal, we select $\mathbf{s}_{i}$ that corresponds to an object whose CLIP feature, $\mathbf{e}_i$, maximizes the dot product with $\mathcal{T}_{\text{target}}$ and $\mathcal{T}_{\text{goal}}$ respectively. These are then passed to the planning module to plan grasping and collision-free motion to fulfill the language command. }
    \label{fig:CLIP_experiment}
\end{figure*}

Algorithm \ref{alg:diffusion_estimator} 
gives a prediction of $\mathcal{S}$, but a higher level of precision is required for complex tasks such as object grasping. To achieve this, we use Algorithm \ref{alg:estimation}, which uses Algorithm~\ref{alg:diffusion_estimator} as a subprocedure to implement an additional refinement step using foreground segmentation. The algorithm begins by initializing an empty set \(\mathcal{C}\) to store the refined $\mathcal{S}$ (L1) and an empty set of evaluations \(\evalsceneset\) to store the evaluation scores of $\mathcal{S}$ (L2). It then predicts foreground segmentation masks \(\{\mathcal{M}_i\}_{i=1}^V\) for each image using a segmentation decoder \(f_{\text{seg}}\) (L3). A simple U-Net architecture is used for \(f_{\text{seg}}\), which is trained with synthetic images and ground truth segmentation labels generated in simulation.  Next, it generates the initial object representations \(\mathcal{S}\) using Algorithm \ref{alg:diffusion_estimator} with a variety of random seeds (L5), and then refines them using gradient descent (L6-8) where the gradient is computed with respect to the function ``Evaluate" (L7). The ``Evaluate" algorithm (Algorithm \ref{alg:evaluatescene}) assesses the quality of $\mathcal{S}$ by comparing the segmentation mask given by the $f_{seg}$ with the ray-hitting results derived from $\mathcal{S}$.  Initially, an empty set \(\evalsceneset\) is created to store the IoU scores for each of \(V\) images (L1, Algorithm~\ref{alg:evaluatescene}). For each image, the algorithm identifies the ray-hitting mask \(\hat{\mathcal{M}}_i\) from \(\mathcal{S}\) for each object using our ray-hitting decoder (L3). The IoU score is then computed by comparing the ray-hitting mask of all objects \(\hat{\mathcal{M}}_i\) with the segmentation mask \(\mathcal{M}_i\) computed by $f_{seg}$ using the ``CalculateIoU'' function (L4). Note that, \(\hat{\mathcal{M}}_i\) and \(\mathcal{M}_i\) consider all objects $\mathcal{S}$, so there is no distinction over objects in these masks. The average IoU score across all images is returned (line 7). After the refinement, the refined  $\mathcal{S}$ is stored in \(\mathcal{C}[b]\) (L9 Algorithm~\ref{alg:estimation}). The scene evaluation score is computed using the ``Evaluate" function and stored in \(\evalsceneset[b]\) (L9-10). Finally, the top $H$ scene representation is selected based on the scene evaluation score from \(\evalsceneset\) (L12). The selected $H$ scene representation \(\{\mathcal{S}_i\}_{i=1}^H\) is returned as the final output.

\subsection{Grounding language to estimated object state representations}
\label{sec:clip_overlay}

Once the set of object states $\{\hat{\objectrepresentation}_i\}_{i=1}^K$ has been estimated, we ground each $\hat{\objectrepresentation}_i$ to a CLIP text embedding.
This process involves two sub-problems: (1) extracting the CLIP features for each object state $\hat{\objectrepresentation}_i$, denoted by $\mathbf{e}_i$ (illustrated 
in Figure \ref{fig:CLIP_experiment}-(a)), and (2) 
using $\mathbf{e}_i$ to identify the specific object referenced in the language command (illustrated in Figure \ref{fig:CLIP_experiment}-(b)).
In the first sub-problem, given a set of observed images $\{I_i\}_{i=1}^V$ with corresponding camera parameters $\{\xi_i\}_{i=1}^V$, and the estimated object representations $\{\hat{\objectrepresentation}_i\}_{i=1}^K$, 
we process the images $\{I_i\}_{i=1}^V$ using a modified CLIP image encoder that omits its final global pooling layer to obtain pixel-wise image embeddings instead of a single vector per image, as in \cite{shen2023distilled}.
To generate corresponding rays for each pixel, we obtain $q_{ray}$ by transforming each pixel into the global positions using the camera's extrinsic parameters. We then establish the ray direction $d_{ray}$ for each pixel based on the camera's intrinsic parameters.
Given these rays, we use \(f_{\text{ray}}\) to determine which pixels correspond to each \(\hat{\objectrepresentation}_i\).
We then aggregate the CLIP features from these identified pixels through mean pooling to create the CLIP feature $\mathbf{e}_i$ for each $\hat{\objectrepresentation}_i$. This step is detailed in Figure\ref{fig:CLIP_experiment}-(a).

Given the set of pairing $(\hat{\objectrepresentation}_i,e_i)^K_{i=1}$,
we use the CLIP text encoder to convert the text of the target and goal objects into their respective CLIP text embeddings.
We then select $\mathbf{s}_{i}$ that corresponds to an object whose $\mathbf{e}_i$ maximizes the dot product with CLIP text embeddings. These are then passed to the planning module to plan grasping and collision-free motion to fulfill the language command. The procedures are described in Figure \ref{fig:CLIP_experiment}-(b).

%% file: experiment_v2.tex
\section{Experiments}

\newcommand{\baselineregression}{\entirename-NonDM}
\newcommand{\baselineimplicit}{\texttt{DEF-}ori\texttt{ON}}
\newcommand{\baselinesingle}{DETR3D-\ourrep}
\newcommand{\baselineimage}{PARQ-\ourrep}
\newcommand{\baselinevoxel}{RayTran-\ourrep}

\newcommand{\baselineshapo}{ShAPO}

\subsection{Simulation experiments} \label{sec:exp_simulation}
We now compare our framework \entirename\ to baselines to validate the two claims: (1) our representation, trained with collision and ray testing decoders, is more computationally efficient than other representations that do not use our decoders and (2) our estimator achieves higher accuracy in estimation and robustness than state-of-the-art baselines.

\begin{figure}[h]
    \centering
    \includegraphics[width=1.0\linewidth]{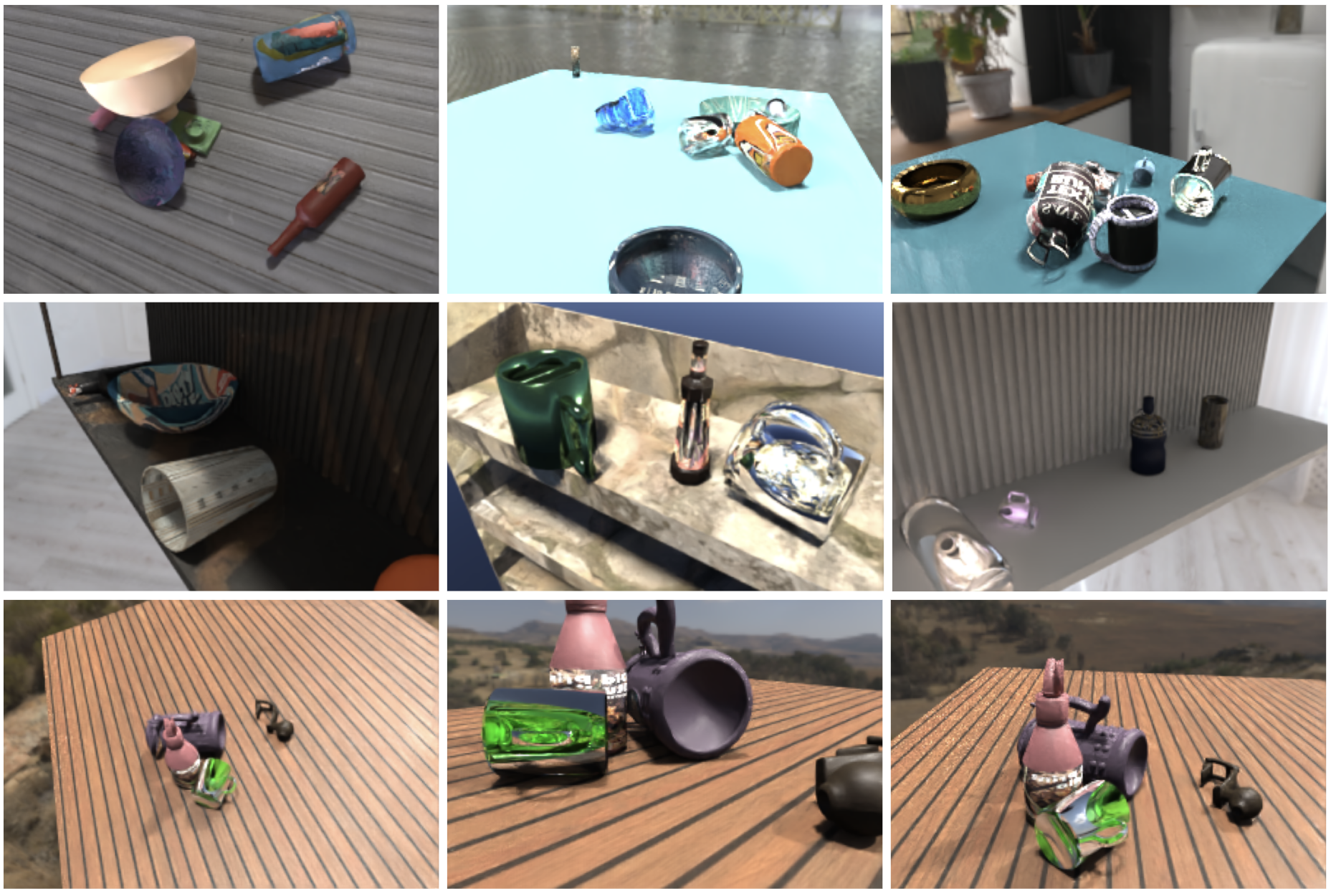}
    \caption{Examples of RGB images from the dataset for diffusion model. The first row shows images from the table environment, while the second row shows images from the shelf environment. Both environments are utilized for training and evaluation.
    The third row shows images from the same data point with different camera parameters.}
    \label{fig:synthetic_data}
\end{figure}

We first compare estimation accuracies in simulated scenes such as shown in Figure \ref{fig:synthetic_data}. We create these scenes by selecting up to 7 objects from an object set and randomly placing them in two types of environments, ``table" and ``shelf." We capture images $\{I_i\}^V_{i=1}$ from 2-5 distinct viewpoints, where camera parameters $\{\camparam_i\}_{i=1}^{\numcam}$ are chosen randomly within a specified boundary.  We use PyBullet~\cite{coumans2021} for physics simulation and a ray-tracing synthetic renderer~\cite{nvisii} for the images. We use the subset of NOCS training object set~\cite{Wang_2019_CVPR_NOCS} for training, which consists of 682 shapes. The training set consists of five categories from ShapeNetCore~\cite{chang2015shapenet}: a bowl, a bottle, a cup, a camera, and a mug. For the table, we use a primitive rectangle shape with random size. For the shelf, we use three shelf shapes from ShapeNet~\cite{chang2015shapenet}. The training dataset consists of 300k scenes. 
Detailed values for these hyperparameters used during experiments can be found in Appendix \ref{app:implementation_details}.

To evaluate the estimation accuracy, we use the same five categories as the NOCS training set but use novel 119 shapes that do not overlap with the training set. We report two quantities: Average Precision (AP) and Chamfer Distance (CD)~\cite{OccupancyNetworks}, where AP evaluates the object detection performance and CD evaluates the accuracy of the predicted shapes. For AP, we follow the evaluation procedure of \cite{irshad2022shapo}: we first sample the point cloud using occupancy decoder $f_{occ}$ (see Appendix \ref{app:pointcloud_sampling} for details) and then extract Oriented Bounding Boxes (OBB) of predicted and ground truth point clouds using Open3D~\cite{zhou2018open3d}. We then calculate the 3D IoU between the ground-truth bounding box and that of the estimated point cloud using PyTorch3D. For symmetric objects, such as bottles, bowls, and cans, we rotate the ground truth bounding box along the symmetry axis and evaluate each rotated instance, then pick the bounding box that maximizes the 3D IoU as described in~\cite{objectron2021}. As done in~\cite{everingham2010pascal}, we first sort the predictions in the descending order of confidence and then match each prediction to one ground truth if the IoU exceeds a threshold. If multiple ground truths have an IoU above the threshold, the prediction is matched to the one with the highest IoU. We do not match a ground truth bounding box if it is already matched. We evaluate AP at IoU25 and 50 thresholds. CD measures the distance between the point clouds of all pairs of matched predicted shapes and ground truth shapes from AP evaluation.

\begin{table*}[t]
    \centering
    \scriptsize 
    \begin{tabular}{lccccc}
        \hline
        \textbf{Methods} & \textbf{@IoU25 AP / CD (mm)} & \textbf{@IoU50 AP / CD (mm)} &  \textbf{Total Estimation Time (s)} & \textbf{Inference Time (s)} & \textbf{Refinement Time (s)} \\
        \hline
        \entirename~(Ours) & \textbf{0.528} / \textbf{0.015} & 0.109 / 0.013 & 0.349 $\pm$ 0.021 & 0.032 $\pm$ 0.001 & 0.317 $\pm$ 0.021 \\
        \baselineregression~(Ours) & 0.524 / \textbf{0.015} & \textbf{0.119} / \textbf{0.012} & \textbf{0.165} $\pm$ 0.022 & 0.010 $\pm$ 0.001 & 0.155 $\pm$ 0.022 \\
        \hline
        \baselinesingle~\cite{detr3d} & 0.279 / 0.018 & 0.012 / 0.015 & 0.329 $\pm$ 0.020 &  0.012 $\pm$ 0.001 & 0.317 $\pm$ 0.020 \\
        \baselineimage~\cite{xie2023pixel} & 0.387 / 0.017 & 0.054 / 0.014 & 0.387 $\pm$ 0.023 & 0.068 $\pm$ 0.009 & 0.320 $\pm$ 0.021 \\
        \baselinevoxel~\cite{raytran} & 0.340 / 0.017 & 0.038 / 0.014 & 0.457 $\pm$ 0.021 & 0.138 $\pm$ 0.001 & 0.320 $\pm$ 0.021 \\
        \hline
    \end{tabular}
    \caption{Results of different methods for estimation in simulation. The values after $\pm$ indicate standard deviation. Inference Time and Refinement Time stand for the time used for inference and refinement in Algorithm \ref{alg:diffusion_estimator} and \ref{alg:estimation}, respectively. We train each model with one seed.}
    \label{tab:estimation_sim}
\end{table*}

We compare \entirename\ against the following models in estimating our representation \ourrep. They use Algorithms \ref{alg:estimation} and \ref{alg:evaluatescene}, and are trained with the same loss function (Equation \eqref{eq:dif_loss}), but use different image feature extractors as proposed in their original papers.
\begin{itemize}
    \item \baselinesingle~\cite{detr3d}: extracts image features based solely on the object's position $c$ rather than using geometric representative points $\mathbf{p}$ like in DETR3D~\cite{detr3d} (Figure \ref{fig:network_comparison}-(c)).
    \item \baselineimage~\cite{xie2023pixel}:
    uses the entire pixel-wise image features (Figure \ref{fig:network_comparison}-(a)).
    \item \baselinevoxel~\cite{raytran}: 
    uses features extracted using a 3D voxel grid. (Figure \ref{fig:network_comparison}-(b)).
\end{itemize}
We use $B=32$ and $H=1$ for the diffusion estimation (Algorithm \ref{alg:estimation}), meaning that the best single prediction among 32 samples is used for estimation evaluation. To evaluate the effect of characterizing uncertainty, we also compare \entirename\ with its variant without a diffusion model:
\begin{itemize}
    \item \baselineregression: 
     \entirename\ trained with regression loss instead of diffusion loss and learns the object queries instead of using Gaussian noise for the initialization. Otherwise the same as \entirename.
\end{itemize}

The results are shown in Table \ref{tab:estimation_sim}. \entirename\ achieves the highest estimation performance among all baselines. Specifically, at IoU25, \entirename\ achieves a 0.141 higher Average Precision (AP) and a Chamfer Distance (CD) 2mm closer to the ground truth shape than the nearest competitor, \baselineimage. Similarly, at IoU50, \entirename\ outperforms \baselineimage\ with a 0.055 higher AP and a 1mm smaller CD.

\begin{figure}[t]
    \centering
    \includegraphics[width=1.0\linewidth]{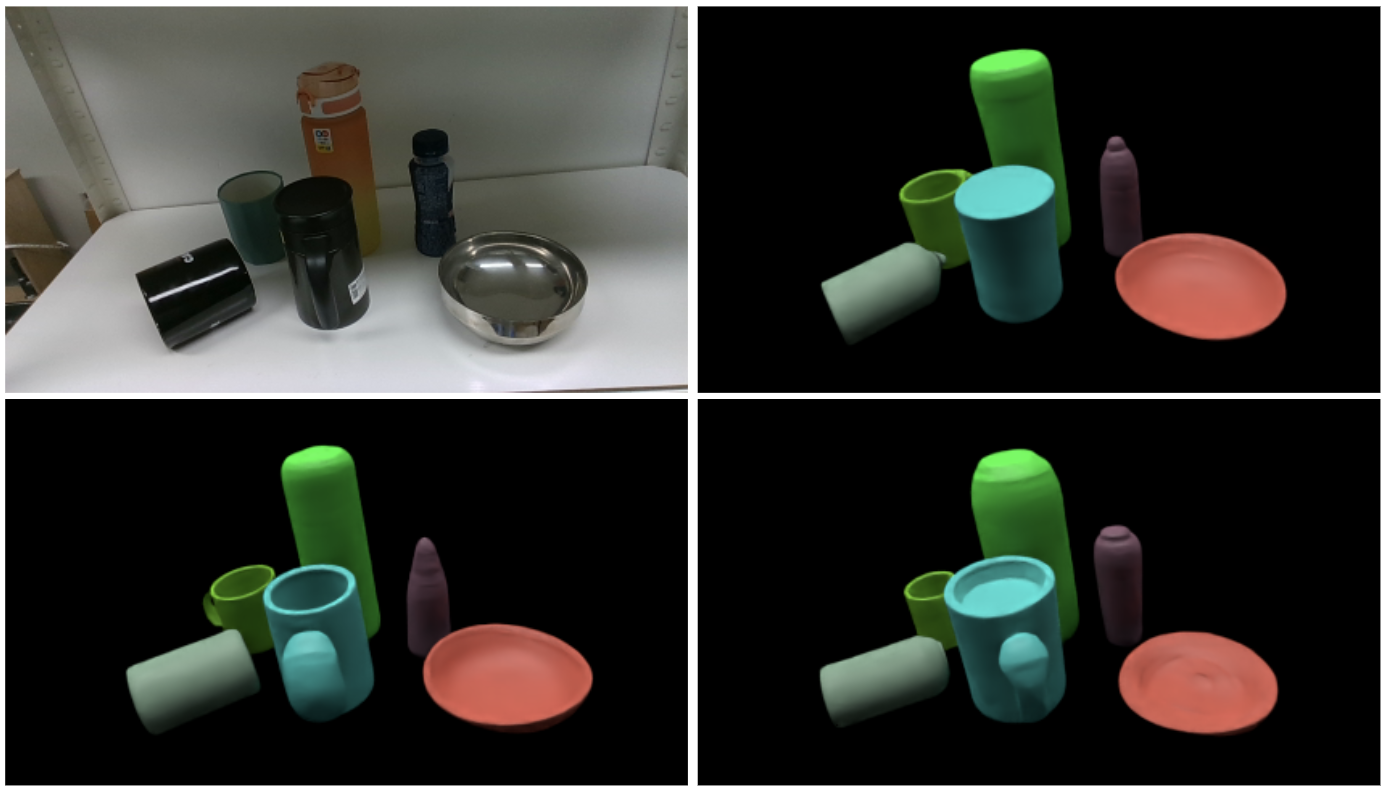}
    \caption{Visualization of stochastic outputs from the diffusion estimator of \entirename\. The upper left figure displays the real-world configuration, while the three other figures show the predictions. The samples capture different details of features that are not clearly distinguishable from the image, such as the handle of the black cup and the shape of the blue cup.}
    \label{fig:estimation_uncertainty}
\end{figure}

While \baselinesingle\ has faster inference speeds than \entirename\ due to a smaller attention computation scope, this comes with a loss in prediction accuracy. Specifically, \baselinesingle\ shows a decrease in AP by 0.249 and 0.097, and an increase in CD by 3mm and 2mm at IoU25 and IoU50, respectively compared to \entirename. \baselineimage\ and \baselinevoxel\ show inference speeds that are two and four times slower than \entirename, respectively, as indicated in the inference time column in Table \ref{tab:estimation_sim}. This is due to the excessive number of features extracted from pixel-wise and voxel-based feature extractors, which result in redundant computations of irrelevant relations, such as empty space. In contrast, using geometric representative points $\mathbf{p}$ in \entirename\ eliminates these redundant computations.

The comparison between \entirename\ and its non-diffusion variant, \baselineregression, shows that both methods achieve nearly identical performance, with 0.004 and 0.010 differences in AP at IoU25 and IoU50, respectively. Despite this similarity in estimation performance, \entirename~has the ability to capture multi-modality, as shown in Figure~\ref{fig:estimation_uncertainty}. This leads to more robust planning, as we show next.


\begin{table*}[t]
    {\footnotesize
    \centering
    \begin{tabular}{lcccc}
        \hline
        \textbf{Methods} & \textbf{Suc. Rate (\%)} & \textbf{No. of Nodes in a Collision} & \textbf{Avg. Max Penetration (mm)} & \textbf{Total Time (s)} \\
        \hline
        \entirename~(Ours)        & \textbf{90.0} & \textbf{1.22} \(\pm\) 5.89 & \textbf{0.84} \(\pm\) 4.47 & 2.626 \(\pm\) 0.105 \\
        \baselineregression~(Ours)  & 87.5          & 2.71 \(\pm\) 10.88         & 1.75 \(\pm\) 7.16          & \textbf{1.845} \(\pm\) 0.106 \\
        \hline
        \baselinesingle~\cite{detr3d} & 71.8          & 6.46 \(\pm\) 15.36         & 4.46 \(\pm\) 10.91         & 2.615 \(\pm\) 0.127 \\
        \baselineimage~\cite{xie2023pixel} & 78.8          & 3.90 \(\pm\) 11.04         & 3.58 \(\pm\) 11.22         & 2.663 \(\pm\) 0.108 \\
        \baselinevoxel~\cite{raytran}      & 75.2          & 4.84 \(\pm\) 12.16         & 3.62 \(\pm\) 10.23         & 2.744 \(\pm\) 0.124 \\
        \hline
        \baselineimplicit           & 73.5          & 2.59 \(\pm\) 5.93 & 1.97 \(\pm\) 5.08 & 154.263 \(\pm\) 8.551 \\
        \hline
    \end{tabular}
    \caption{Evaluation of Methods for motion planning in simulation. The values after $\pm$ indicate standard deviation. Suc. Rate, No. of Nodes in a Collision, and Avg. of Max Penetration. We report the numbers from 3 repeated evaluations of 200 scenes with three different random seeds.}
    \label{tab:motion_planning_sim}
    }
\end{table*}

\begin{figure}[t]
    \centering
    \includegraphics[width=0.75\linewidth]{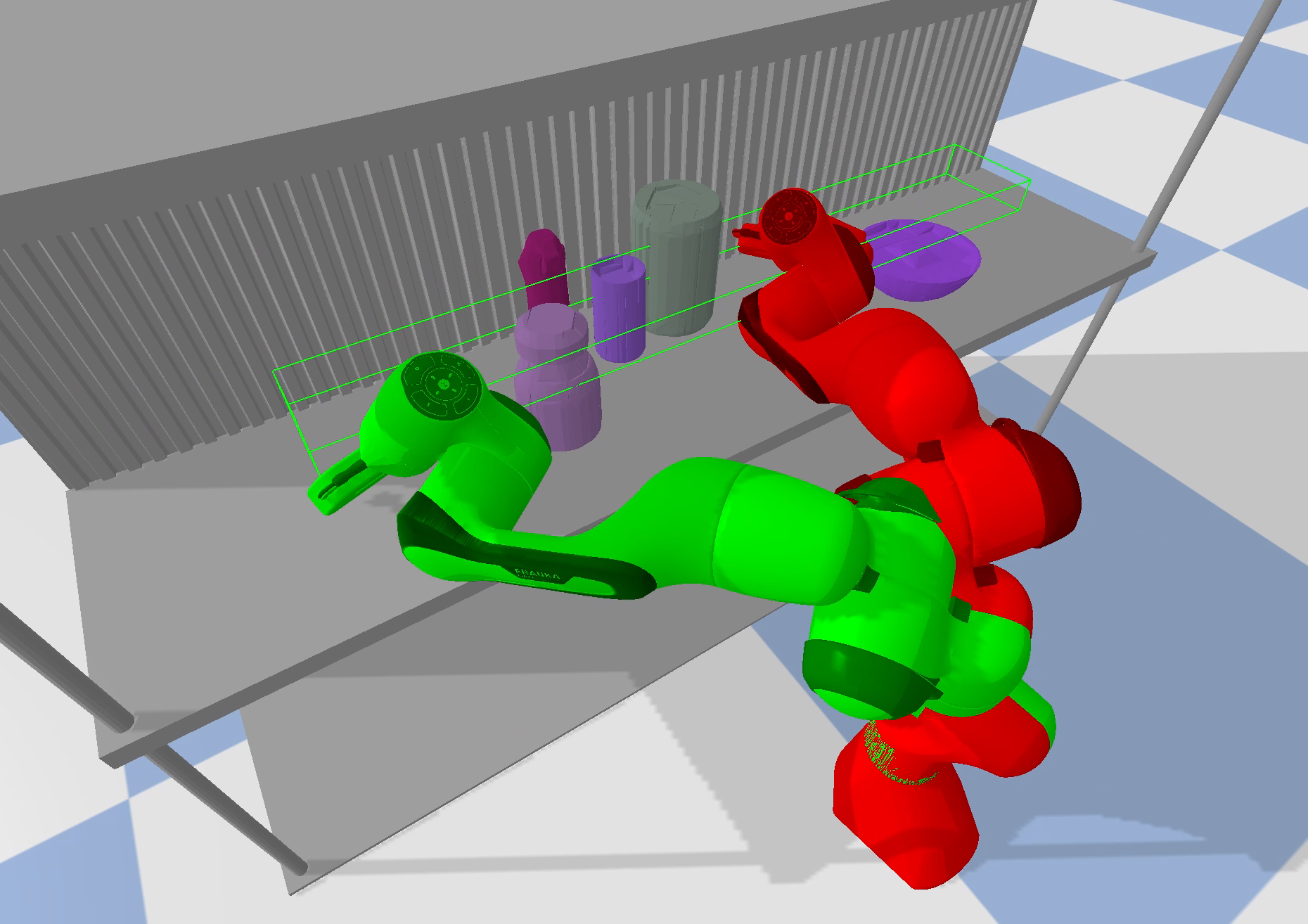}
    \caption{An example of our motion planning problem. The start and goal configurations of the robot are highlighted in red and green, respectively. We randomly select objects and randomly choose their locations and the shape of the shelf.}
    \label{fig:exp_cluttered_env_sim}
\end{figure}

We evaluate our framework in collision-free motion planning problems, where the objective is to plan a collision-free path for Franka Panda in a tight environment using different representations and estimators. The setup for this task is shown in Figure \ref{fig:exp_cluttered_env_sim}. We capture three images using NVISII renderer~\cite{nvisii} for estimation. For motion planning using our representation \ourrep, we use $f_{col}$ between the object and robot for collision prediction and modify RRT*~\cite{rrtstar} to account for uncertainty (see Appendix \ref{app:planning_with_uncertainty}). We use $H=2$ and $B=32$ for the diffusion estimator (Algorithm \ref{alg:estimation}), meaning that the two best predictions among 32 samples are used for the motion planning.

To evaluate the efficiency of our representation \ourrep, we compare against an alternative representation trained without collision decoder $f_{col}$ and ray testing decoder $f_{ray}$, which we call \textbf{ori}ented \textbf{O}ccupancy-based representatio\textbf{N} (ori\texttt{ON}). When combined with our diffusion-based estimator, DEF, this forms the baseline algorithm \baselineimplicit.
To check collision for this baseline, we use the following scheme~\cite{le2023differentiable}, using $f_{occ}$: first sample the surface point cloud of a pair of objects using the occupancy decoder $f_{occ}$,  and declare collision if any point of one object is inside the other. To see the impact of estimation performance on motion planning, we also compare \entirename\ against the same set of estimators from the estimation experiment. 

We measure the ``success rate," ``the number of nodes in a collision on a trajectory," and ``the average max penetration depth on a trajectory." A trajectory is represented as 200 equidistant nodes. Each node undergoes collision detection using the ground truth object shapes in PyBullet. A trajectory is a success if none of the nodes is in collision. We evaluate methods in 200 scenes, averaged across 3 repeated evaluations with different random seeds.

\begin{figure}[h]
    \centering
    \includegraphics[width=1.0\linewidth]{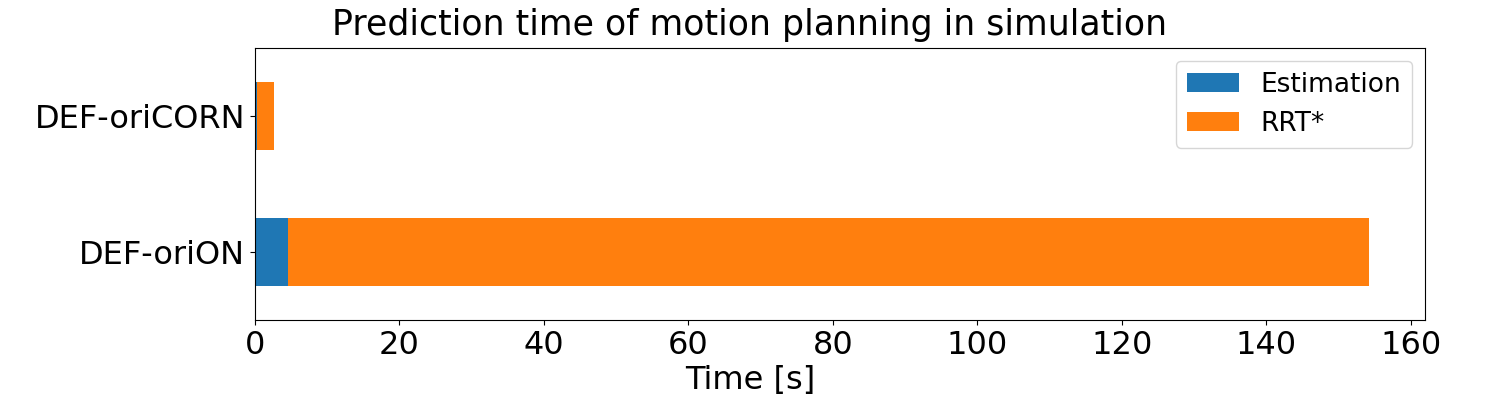}
    \caption{Estimation and planning times for the motion planning problem.}
    \label{fig:motion_planning_bar_graph}
\end{figure} 

The results are shown in Table \ref{tab:motion_planning_sim}. We can see the impact of estimation accuracy on motion planning: \entirename\, which achieved the highest estimation accuracy, achieves the highest success rate. To investigate the impact of introducing the uncertainty, we compare \entirename\ with its non-diffusion variant. \entirename\ achieves a 2.5\% higher success rate than \baselineregression, with 1.49 fewer nodes in collision and a 0.91mm smaller average maximum penetration depth. This indicates that \entirename\ gives more robust plans, with fewer nodes in deep penetration collisions, showing the effect of taking uncertainty into account. Figure \ref{fig:estimation_uncertainty} qualitatively shows what stochasticity in our diffusion model can capture.

In terms of the computational efficiency of \ourrep\ in comparison to \baselineimplicit~that only has an occupancy decoder, all methods with \ourrep\ take less than 3 seconds to complete from estimation to motion planning, while \baselineimplicit\ requires 150 seconds, making it more than 50 times slower. 
This is because \entirename\ uses $f_{ray}$ and $f_{col}$ for estimation refinement and collision detection, whereas \baselineimplicit\ has to excessive number of evaluations of $f_{occ}$ for collision detection and ray-hitting test. More concretely, 
for collision detection under \baselineimplicit, the surface point cloud for a pair of objects is sampled using $f_{occ}$, as detailed in Appendix \ref{app:pointcloud_sampling}, with a collision declared if any point of one object penetrates the other.
Similarly, the ray-hitting test involves sampling points along a ray, declared through simple interpolation with a pre-defined resolution, and identifying ray-hitting when any point resides inside the object.
In both tests, $f_{occ}$ must be evaluated numerous times, whereas \entirename\ only requires a single prediction from $f_{col}$ and $f_{ray}$, making \baselineimplicit\ significantly slower than \entirename.
The breakdown of the computation time is shown in Figure \ref{fig:motion_planning_bar_graph}.

Additionally, \baselineimplicit\ has the second-lowest success rate. 
This is because we have to restrict the number of evaluations of $f_{occ}$ for collision and ray-hitting tests due to constraints on GPU memory and processing time. 
Such limitations directly compromise the accuracy of collision and ray-hitting detection, thereby decreasing the overall planning performance.
Unlike \baselineimplicit, methods employing \ourrep\ decoders avoid these issues as they do not rely on extensive point sampling for collision and ray-hitting tests, thanks to $f_{col}$ and $f_{ray}$.

\subsection{Real-world experiments} \label{sec:exp_real}

\begin{figure}[t]
    \centering
    \includegraphics[width=1.0\linewidth]{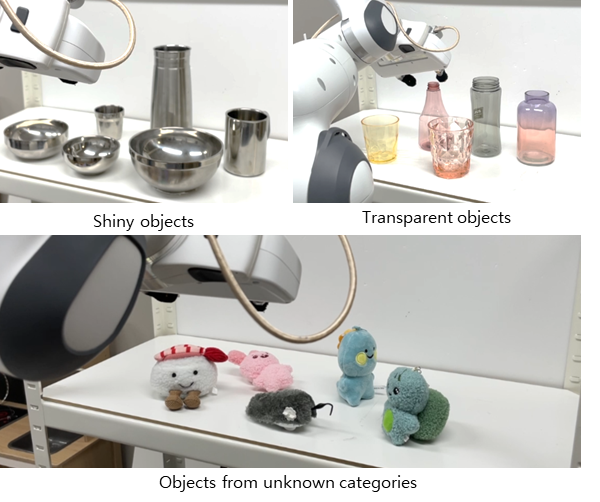}
    \caption{Examples of objects used for the real-world experiments. It includes transparent objects (top left) and shiny objects (top right) that are difficult to detect with depth sensors. 
    Below, we illustrate our system's ability to generalize to unknown objects, which are distinct in geometry from those encountered during training (such as cups, mugs, bowls, bottles, and cans).
    }
    \label{fig:real_objects}
\end{figure}

\begin{figure*}[t]
    \centering
    \includegraphics[width=1.0\linewidth]{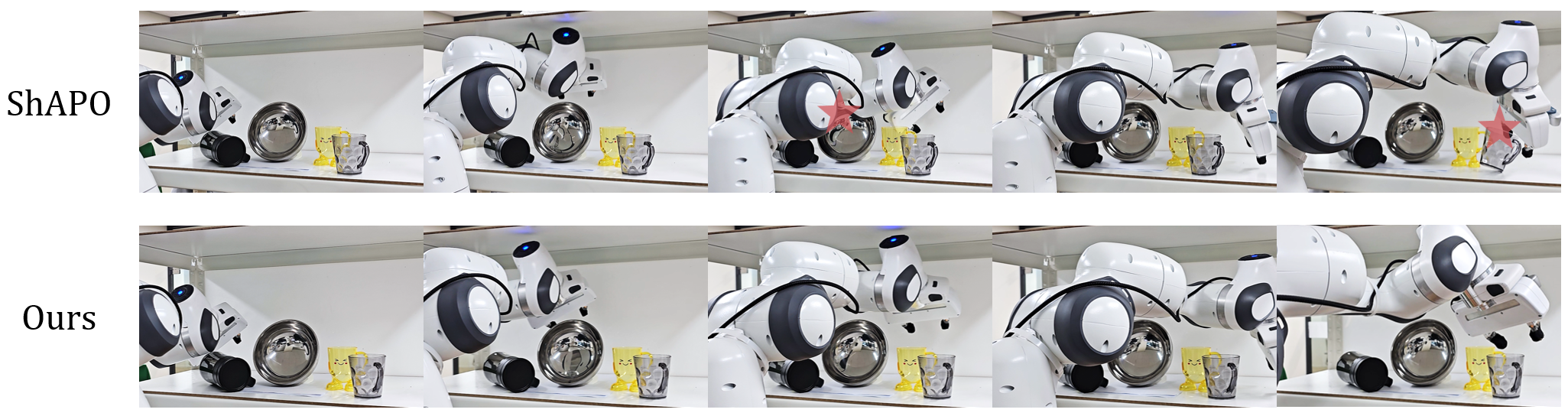}
    \caption{Qualitative visualization of the failure modes of \baselineshapo\ with transparent/shiny objects in motion planning. The collisions are highlighted with a red star.}
    \label{fig:motion_planning_ours_shapo}
\end{figure*}

In this section, we compare different methods in real-world applications where we have novel and diverse objects with different materials that were never seen during training. Some example objects we use are shown in Figure~\ref{fig:real_objects}. We first compare baselines in real-world motion planning, as shown in Figure \ref{fig:motion_planning_ours_shapo}. The robot's substantial volume makes it a narrow passage problem, which is especially difficult for sampling-based motion planners.

\begin{figure*}[t]
    \centering
    \includegraphics[width=1.0\linewidth]{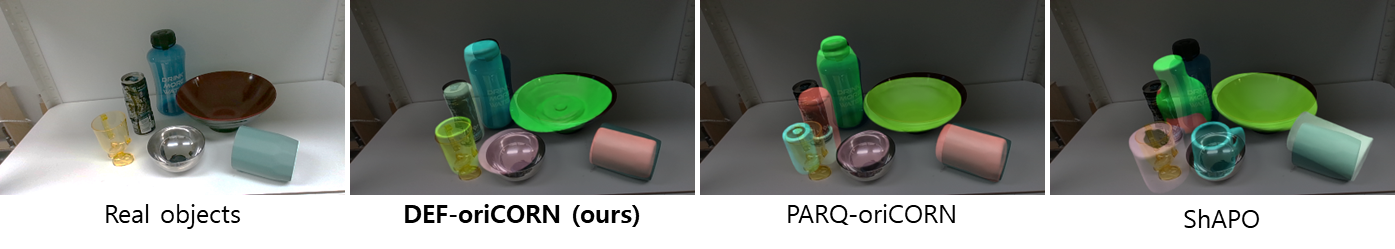}
    \caption{The left figure shows an example of a real-world scene. \entirename\ predicts all objects correctly with accurate shape boundaries. \baselineimage\, which has the second-best quantitative results, contains false positives in its predictions, and the shape boundaries do not coincide with the real objects. \baselineshapo\ fails to estimate the shape of the shiny bowl, transparent blue bottle, and yellow cup.}
    \label{fig:estimation_quality}
\end{figure*}

We have two baselines. First, to evaluate the efficiency of our representation \ourrep\ and the effectiveness of using RGB images, we compare \entirename\ against \baselineshapo, that uses implicit shape (an SDF) and 6D pose of objects and depth images. Second, to see the impact of estimation performance on motion planning in the real world, we compare \entirename\ against \baselineimage, which showed the best overall performance among different estimators in the simulation experiments. We use a reconstructed mesh using Marching Cubes~\cite{lorensen1998marching} for collision detection for \baselineshapo. For details as to how we trained \baselineshapo, see Appendix \ref{app:shapo}.

We measure the success rate to evaluate where success is defined as no collision between the robot arm and scene when a planned trajectory is executed. The results are in Table~\ref{tab:motion_planning_real}. We see that \entirename\ shows a 22\% higher success rate than \baselineimage\ in the real world. \baselineshapo\ has a zero success rate, primarily due to collisions with transparent and shiny items. We show the qualitative estimation in Figure~\ref{fig:estimation_quality}. We observe that \baselineshapo\ misses or incorrectly predicts transparent objects and \baselineimage\ often struggles with precise alignment of object locations relative to actual observations.

To compare the efficiency of different representations, we also compare the time taken for estimation and motion planning. \baselineimage\ is 0.03 seconds slower than \entirename\ due to the redundant computation in its architecture, consistent with the simulation result. While the methods with our representation \ourrep take less than 3 seconds in total, \baselineshapo\ requires more than 100 seconds. This is because of the occupancy-based shape optimization of \baselineshapo\ during estimation and Marching Cubes~\cite{lorensen1998marching} for mesh reconstruction. 

\begin{table}[h]
\small
\centering
    \begin{tabular}{lcc}
    \toprule
    \textbf{Methods}      & \textbf{Success Rate} & \textbf{Time (s)} \\
    \midrule
    \entirename~(Ours)  & \textbf{9/10} & \textbf{2.605} \(\pm\) 0.114 \\
    \baselineimage~\cite{xie2023pixel}        & 7/10 & 2.639 \(\pm\) 0.131 \\
    \baselineshapo~\cite{irshad2022shapo}        & 0/10 & 102.995 $\pm$ 13.199  \\
    \bottomrule
    \end{tabular}
    \caption{Evaluation of methods on success rate and computation time for motion planning in the real world. The values after $\pm$ indicate standard deviation.}
\label{tab:motion_planning_real}
\end{table}

Next, we evaluate the baselines in a language-commanded pick-and-place problem. In this task, a robot retrieves a designated object from a shelf and places it at a target location, following the instructions provided in the language. The language command has templates as follows:
\begin{itemize}
\item \textbf{Fetch / Give me} [\texttt{obj}]
\item \textbf{Place / Move} [\texttt{obj1}] [\texttt{spatial\_loc}] of [\texttt{obj2}]
\end{itemize}
where \texttt{spatial\_loc} can be any of \{``\textit{left}," ``\textit{right}," ``\textit{in front}," ``\textit{behind}," ``\textit{inside}," ``\textit{up}"\}. The real-world setup is illustrated in Figure \ref{fig:problem_setup}.

For grasping, we begin by sampling potential 6D grasp poses within a specified bounded box. The feasibility of these poses is checked using $f_{col}$ to ensure there is no collision between the robot's gripper and any surrounding objects. Validated grasps are further evaluated based on the antipodal heuristic~\cite{nguyen1988constructing, modernrobotics}, focusing on the grip's stability and viability.
This evaluation requires identifying contact points and their corresponding normal vectors, obtained by interpolating points along the line connecting the gripper tips and assessing these points with the occupancy predictor $f_{occ}$. The gradients of $f_{occ}$ at these points provide the necessary normal vectors.
To assess the stability of a grasp, we calculate the minimum friction cone angles at the contact points, aligned with their corresponding contact normals, ensuring that a line connecting these points remains within both cones for a more stable grasp.
The most suitable grasp is selected based on these criteria. We also conduct an experiment comparing different estimators and representations for grasping but omit it here for brevity. Interested readers should see Appendix \ref{app:exp_grasping_sim}.

We measure the success rate and time, where a trial is a success if the robot achieves the language command.
The success rates are shown in Table \ref{tab:pnp_real}. \entirename\ achieves a 75\% success rate, which shows that our estimation and representation \ourrep\ can indeed ground objects to achieve language-aligned representations.  

\begin{table}[h]
    \small
    \centering
    \begin{tabular}{lccccc}
        \hline
        \textbf{Category} & \textbf{Bottle} & \textbf{Cup} & \textbf{Can} & \textbf{Bowl} & \textbf{Total}\\
        \hline
        Suc. Rate & 4/5 & 4/5 & 3/5 & 4/5 & 15/20 (75\%) \\
        \hline
    \end{tabular}
    \caption{Evaluation of methods on success rate for language-guided pick-and-place in real world}
    \label{tab:pnp_real}
\end{table}

We also investigate the computational efficiency in grounding language to objects using different representations, specifically comparing \entirename\ and \baselineimplicit. To ground text to object in \baselineimplicit\, we follow the exact same procedure as our method, but instead of using $f_{ray}$, we do ray-testing using $f_{occ}$ by sampling points along a ray, and assessing them with $f_{occ}$ to check if the ray hits the object. As shown in Table \ref{tab:CLIP_grounding_time}, the time taken for language grounding by \entirename\ is about three times faster than \baselineimplicit.  This is because the ray-hitting test of \entirename\ utilizes $f_{ray}$ to predict ray interactions through a single evaluation, significantly reducing the computational demand compared to the multiple evaluations required by \baselineimplicit.

\begin{table}[h]
    \centering
    \begin{tabular}{lc}
        \hline
        \textbf{Methods}      & \textbf{Time (s)}\\
        \hline
        \entirename~(Ours)  & \textbf{0.088} $\pm$ 0.002\\
        \baselineimplicit     & 0.250 $\pm$ 0.002 \\
        \hline
    \end{tabular}
    \caption{Evaluation of methods on computational efficiency for CLIP-based language grounding}
    \label{tab:CLIP_grounding_time}
\end{table}


%% file: appendix.tex
\appendix

\section{Planning with uncertainty} \label{app:planning_with_uncertainty}

\subsection{Algorithms}
One strength of our approach is its ability to express uncertainty about estimation results using a diffusion estimator (Figure \ref{fig:estimation_uncertainty}). Specifically, we employ a sampling-based approximation to approximate the distribution of object representations. To achieve this, given a large $B$, we take top-H samples of $\mathcal{S}$ from diffusion estimator(Algorithm \ref{alg:estimation}) instead of only taking the top-1 sample. To account for the uncertainty represented in $H$ samples of $\mathcal{S}$, we apply minor modifications to standard tools such as RRT~\cite{rrt, rrtstar} and grasping heuristic score~\cite{modernrobotics, doi:10.1177/027836498800700301, fang2020graspnet} to incorporate uncertainty in the object representations. 

For motion planning, we use RRT*~\cite{rrtstar} but check the validity of the path by contact test using $H$ samples of $\mathcal{S}$. This ensures that the planned motion accounts for the uncertainty in the object representations, leading to higher performance, as shown in the experiment (Section \ref{sec:exp_simulation}).

For grasping planning, given $H$ number of object representation samples of the target object, we uniformly sample grasp candidates within a bounded range and evaluate them using an antipodal-based heuristic score~\cite{modernrobotics, doi:10.1177/027836498800700301, fang2020graspnet}. This results in $H$ grasp scores for each object sample, which are then averaged to obtain the final grasp score.

To utilize \ourrep\ for grasp planning, we randomly sample 6D end-effector poses near the object and select the best grasp pose using the grasping heuristic score~\citep{modernrobotics, doi:10.1177/027836498800700301, fang2020graspnet}, which is the function of two contact points and their contact normals. Given 37500 random samples near the object, we first filter out grasp poses that collide with the object using $f_{col}$. To compute the heuristic score~\cite{modernrobotics, doi:10.1177/027836498800700301, fang2020graspnet} for each remaining 6D end-effector pose sample, we use $f_{occ}$ to locate the contact point and predict the contact normal of grasping. We evaluate 80 evenly spaced points along the line segment connecting the two gripper tip ends. To detect the contact point, we select the two points closest to each gripper tip that have occupancy values inside the shape. We then determine the contact normal by calculating the gradient of the occupancy at the contact point.

\subsection{Grasping in a tabletop environment} \label{app:exp_grasping_sim}

\subsubsection{Task setup} 
\begin{figure}[h]
    \centering
    \includegraphics[width=0.75\linewidth]{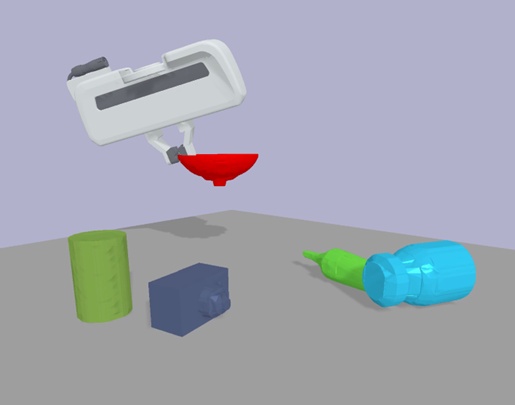}
    \caption{An example of a scenario for grasping. The target object is highlighted in red color. Franka Panda gripper successfully holds and lifts the object the target object without dropping it.}
    \label{fig:grasp_env_sim}
\end{figure}

We create a tabletop scenario with five randomly chosen objects from the NOCS validation set. These objects are arbitrarily positioned on a table, with one random object selected as the grasp target object. We use 200 scenarios for evaluation. Figure \ref{fig:grasp_env_sim} illustrates an example scenario for grasping. 

Similar to the camera setup in the motion planning experiment, we place three cameras on one side of the table and capture images using NVISII renderer~\cite{nvisii}. Then, we estimate $\objectrepresentation$ and plan the grasp pose for the target object. We use $\objectrepresentation$ whose position $c$ is closest to the position of the grasp target object for the grasp planning.

We use $H=2$ and $B=32$ for the diffusion estimator (Algorithm \ref{alg:estimation}), meaning that the two best predictions among 32 samples are used for the grasp planning in consideration of the uncertainty in the estimation.

\subsubsection{Metric} 
The evaluation metric is a success rate of 200 attempts averaged across 3 repeated evaluations with different random seeds. A grasp is determined successful if the gripper securely holds and lifts the object to a height of 50 cm without dropping it.

\subsubsection{Results} 
The results are presented in Table \ref{tab:grasping_sim}. Following previous experiments, we investigate the impact of estimation performance on downstream tasks. \entirename\ shows a 3.9\% higher success rate than the best-performing baseline (\baselineimage). 
The low estimation accuracy of \baselinesingle\ led to significantly lower manipulation performance, with a 12\% lower success rate than \entirename.

To investigate the impact of introducing the uncertainty of the diffusion estimator on grasping performance, we compare \entirename\ with the non-diffusion variant. \entirename\ exhibits a 4.4\% higher success rate than \baselineregression, consistent with the estimation results from Section \ref{sec:exp_simulation}

\begin{figure}[h]
    \centering
    \includegraphics[width=1.0\linewidth]{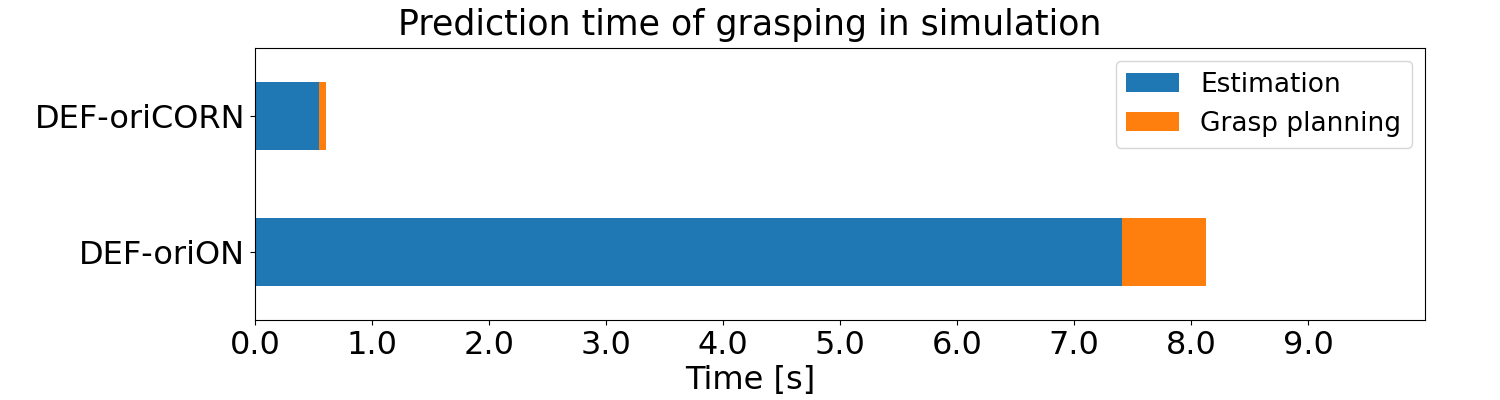}
    \caption{Prediction time of grasping in the simulation environment. The bar chart illustrates the time components in grasp planning. The chart breaks down the total time into phases: Estimation and grasp planning.}
    \label{fig:grasp_bar_chart}
\end{figure}

We compare the computational efficiency of our representation to that of the occupancy-based implicit representation variant. While all methods that use our representation (with \ourrep- prefix) take less than 1 second to complete from estimation to grasping, \baselineimplicit\ requires 8 seconds. The difference in time is due to the usage of $f_{ray}$ and $f_{col}$ for \entirename, as opposed to $f_{occ}$ being used repeatedly to evaluate the query points on the rays from each pixel in the image for ray-hitting test in the refinement and to evaluate the occupancy of surface point clouds~\cite{xu2021end} for collision detection for \baselineimplicit, as illustrated in Figure \ref{fig:grasp_bar_chart}.

\begin{table}[h]
    \footnotesize
    \centering
    \begin{tabular}{lcc}
        \hline
        \textbf{Methods}           & \textbf{Suc. Rate (\%)} & \textbf{Time (s)} \\
        \hline
        \entirename~(Ours)       & \textbf{89.2} & 0.604 \(\pm\) 0.018 \\
        \baselineregression~(Ours) & 84.8 & \textbf{0.239} \(\pm\) 0.041 \\
        \hline
        \baselinesingle            & 77.2 & 0.549 \(\pm\) 0.019 \\
        \baselineimage             & 85.3 & 0.708 \(\pm\) 0.018 \\
        \baselinevoxel             & 82.3 & 0.910 \(\pm\) 0.014 \\
        \hline
        \baselineimplicit & 89.2 & 8.123 \(\pm\) 0.027 \\
        \hline
    \end{tabular}
    \caption{Evaluation of methods on success rate and computation time for grasping in simulation. The values after $\pm$ indicate standard deviation.}
    \label{tab:grasping_sim}
\end{table}

\section{Implementation details}
\label{app:implementation_details}

In our estimation models, including the proposed method, we set the number of diffusion timesteps, $T$, to 5 to achieve a balance between efficiency and accuracy. The decision for this configuration is backed by an ablation study detailed in Appendix \ref{app:ablation_num_inf_step}.
For the refinement step in Algorithm \ref{alg:estimation} (line 6-8), 
the optimization process is performed using the Adam algorithm~\cite{kingma2014adam} with Autograd function in Jax~\cite{jax2018github}, configured with a learning rate of 0.004. Typically, 100 iterations suffice for the refinement process. 

For $f_{enc}$, we use FER-VN-OccNet encoder~\cite{son2024an} with a linear layer width of 64 and with the feature dimension of 3+5. For $f_{occ}$, we use FER-VN-OccNet decoder with a linear layer width of 32. For the $f_{col}$, we first apply pre-processing described in Section \ref{sec:construct_geometric_objrep}, and concatenate $c$ and $z$ after 2 layers of Multilayer Perceptron (MLP), followed by 3 layers and 128 widths of MLP to output collision results.

For $f_{den}$, we use $M=32$ for $\mathbf{p}$, $L=2$, number of head as 4, size of key-query-value in attention layers are 32. We apply AdaNorm~\cite{dhariwal2021diffusionbeat} after each attention. 
For the head for $c, \mathbf{p}, z$ and $o_{conf}$, we use 2 layers and 128 widths of MLP.
For the image encoder, we use U-Net~\cite{ronneberger2015unet} with 3 hierarchies and ray positional embedding~\cite{liu2022petr}.

For the training, we use 10 for $\lambda_{occ}$, 1 for $\lambda_{col}$ and $\lambda_{ray}$, 4 for $h$, 90 for $\alpha_p$, 200 for $\alpha_c$, and 1 for $\alpha_z$. We use Adam optimizer with a learning rate of 0.0004 for both representation pre-training and estimator learning.

\section{Implementation of baselines}

\subsection{ShAPO}
\label{app:shapo}

We trained ShAPO using the same estimation dataset as our method, as described in Section \ref{sec:exp_simulation}. Since our method is trained with multi-view RGB images, we included additional depth images for ShAPO's training. Each sample in our dataset comprises five different views, and we used all five views for training, resulting in a total of 1,500,000 samples for ShAPO. To address the simulation-to-reality gap, especially regarding depth image noise, we incorporated simulated depth noise according to ShAPO's official implementation. We used the SDF representation pre-trained with the NOCS dataset provided with ShAPO's official code.

Motion planning with ShAPO employed the PyBullet simulator~\cite{coumans2021} for collision assessments, reconstructing meshes via marching cubes~\cite{lorensen1998marching} and accelerating the contact detection with vhacd~\cite{mamou2016vhacd} for RRT* efficiency.

\section{Algorithm for point cloud sampling} \label{app:pointcloud_sampling}
In order to sample the surface point cloud using occupancy predictor $f_{occ}$, we utilize a hierarchical sampling technique similar to that described in ShAPO~\cite{irshad2022shapo}. This technique includes the following steps. Initially, grid points within a predefined box at a coarse resolution are evaluated with the occupancy decoder. Points near surfaces, identified by a predefined occupancy value threshold, are gathered and then resampled at a finer resolution to refine the surface points. Finally, we project points onto the surface by estimating the surface normals with gradient descent.

\section{Ablation study on the number of timesteps in the diffusion process}
\label{app:ablation_num_inf_step}

\begin{figure}[t]
    \centering
    \includegraphics[width=1.0\linewidth]{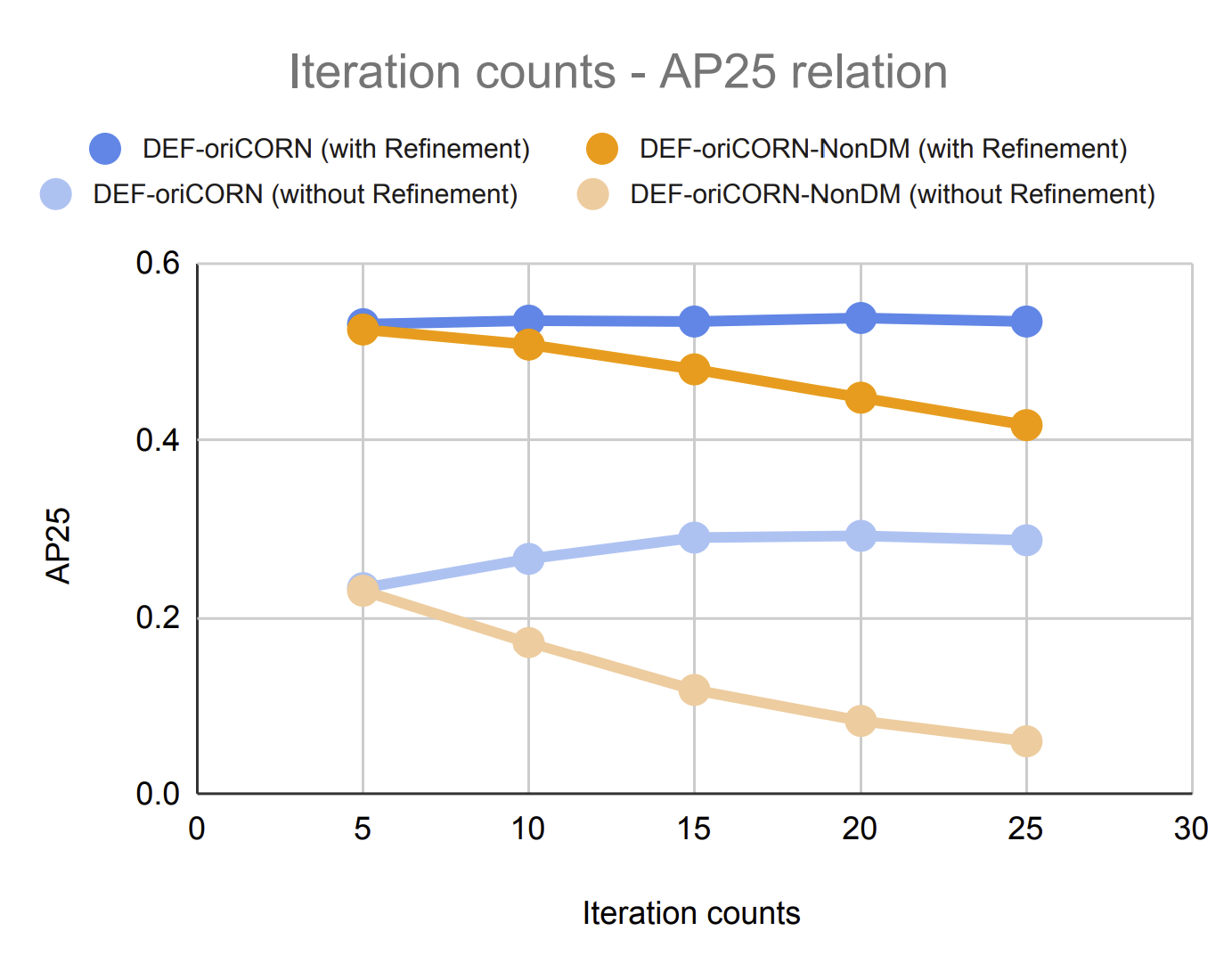}
    \caption{This graph displays the estimation module's performance as a function of iteration number changes in the diffusion process. The x-axis represents different iteration counts, while the y-axis indicates the average precision (AP). The blue curves represent the performance with the diffusion model(\entirename), whereas the orange curves demonstrate the performance without the diffusion model(\baselineregression).}
    \label{fig:ablation_itr}
\end{figure}

A distinctive feature of diffusion models is their capacity to balance accuracy and efficiency by adjusting the number of time steps involved in the diffusion process, as highlighted by \cite{song2020denoising}. To examine this dynamic and contrast it with a regression model baseline (\baselineregression), we conducted an experiment to analyze estimation performance across varying numbers of recurrent iterations (or time steps in the diffusion process). The findings are illustrated in Figure \ref{fig:ablation_itr}.

An insight from our analysis is that an increase in recurrent iterations within the \baselineregression\ tends to deteriorate the estimation model's performance. This decline could be attributed to accumulating errors through recurrent updates, causing the model to drift from the training distribution as iterations advance.

Conversely, the proposed method, which introduces a diffusion model training scheme, exhibits improved performance with increased time steps. This improvement is likely due to the introduction of noise during recurrent processing in training, which enhances the model's resilience against update-induced errors. This attribute allows for enhanced accuracy through augmented iteration steps, demonstrating the diffusion model's capability to rectify issues prevalent in previous methodologies.